%% file: main.tex
\documentclass[11pt, a4paper, logo, copyright, nonumbering]{deepseek}
\usepackage[authoryear, sort&compress, round]{natbib}
\usepackage{dblfloatfix}
\usepackage{ulem}
\usepackage{caption}
\usepackage{dramatist}
\usepackage{xspace}
\usepackage{pifont}% http://ctan.org/pkg/pifont
\usepackage{multirow}
\usepackage{adjustbox}
\usepackage{tcolorbox}
\usepackage{xltabular}
\usepackage{longtable}
\usepackage{hyperref}
\usepackage{makecell}
\usepackage{algorithm}
\usepackage[noend]{algpseudocode}

\usepackage{amsfonts}
\usepackage{amsmath}
\usepackage{amssymb}
\usepackage{lineno}

\usepackage[bottom]{footmisc}
\usepackage{CJKutf8}
\usepackage{subfigure}
\usepackage{setspace}

\usepackage{titlesec}
\usepackage{enumitem}
\makeatletter
\def\@BTrule[#1]{%
  \ifx\longtable\undefined
    \let\@BTswitch\@BTnormal
  \else\ifx\hline\LT@hline
    \nobreak
    \let\@BTswitch\@BLTrule
  \else
     \let\@BTswitch\@BTnormal
  \fi\fi
  \global\@thisrulewidth=#1\relax
  \ifnum\@thisruleclass=\tw@\vskip\@aboverulesep\else
  \ifnum\@lastruleclass=\z@\vskip\@aboverulesep\else
  \ifnum\@lastruleclass=\@ne\vskip\doublerulesep\fi\fi\fi
  \@BTswitch}
\makeatother

\addto\extrasenglish{
}

 {\begin{list}{}%
         {\setlength{\leftmargin}{#1}}%
         \item[]%
 }
 {\end{list}}
 
\reportnumber{001} % Leave blank if n/a

\renewcommand{\today}{}
\newcommand{\spmath}{DeepSeekMath}

\title{\centering \spmath: Pushing the Limits of Mathematical Reasoning in Open Language Models}
\vspace{-5mm}
\author[*]{
\small
\hspace{2em}
Zhihong Shao$^{1,2*\dag}$, Peiyi Wang$^{1,3*\dag}$, Qihao Zhu$^{1,3*\dag}$, Runxin Xu$^{1}$, Junxiao Song$^{1}$
\newline
Xiao Bi$^{1}$,
Haowei Zhang$^{1}$,
Mingchuan Zhang$^{1}$,
Y.K. Li$^{1}$, 
Y. Wu$^{1}$,
Daya Guo$^{1*}$

\small
$^1$DeepSeek-AI, $^2$Tsinghua University, $^3$Peking University \\
\small
\texttt{\{zhihongshao,wangpeiyi,zhuqh,guoday\}@deepseek.com} \\
\small
\url{https://github.com/deepseek-ai/DeepSeek-Math}
}
\correspondingauthor{\, $^*$ Core contributors. \\\, $^\dag$ Work done during internship at DeepSeek-AI.}

\input{commands.tex}

\begin{abstract}
Mathematical reasoning poses a significant challenge for language models due to its complex and structured nature. %On the MATH benchmark, the best-open-sourced model just reached 30$\%$+ top-1 accuracy, while closed models of big companies have surpassed the 40\% milestone. 
In this paper, we introduce \spmath~7B, which continues pre-training DeepSeek-Coder-Base-v1.5 7B  with 120B math-related tokens sourced from Common Crawl, together with natural language and code data. \spmath~7B has achieved an impressive score of 51.7\% on the competition-level MATH benchmark without relying on external toolkits and voting techniques, approaching the performance level of Gemini-Ultra and GPT-4.  Self-consistency over 64 samples from \spmath~7B achieves 60.9\% on MATH. %Furthermore, it demonstrates remarkable proficiency on other mathematical reasoning benchmarks such as GSM8K, SAT, C-Math, and Gaokao.
The mathematical reasoning capability of \spmath~is attributed to two key factors: First, we harness the significant potential of publicly available web data through a meticulously engineered data selection pipeline. Second, we introduce Group Relative Policy Optimization (GRPO), a variant of Proximal Policy Optimization (PPO), that enhances mathematical reasoning abilities while concurrently optimizing the memory usage of PPO.
 \end{abstract}

\begin{document}
\begin{CJK*}{UTF8}{gbsn}

\maketitle

\begin{figure}[h]
\begin{center}
        \includegraphics[width=0.68\textwidth]{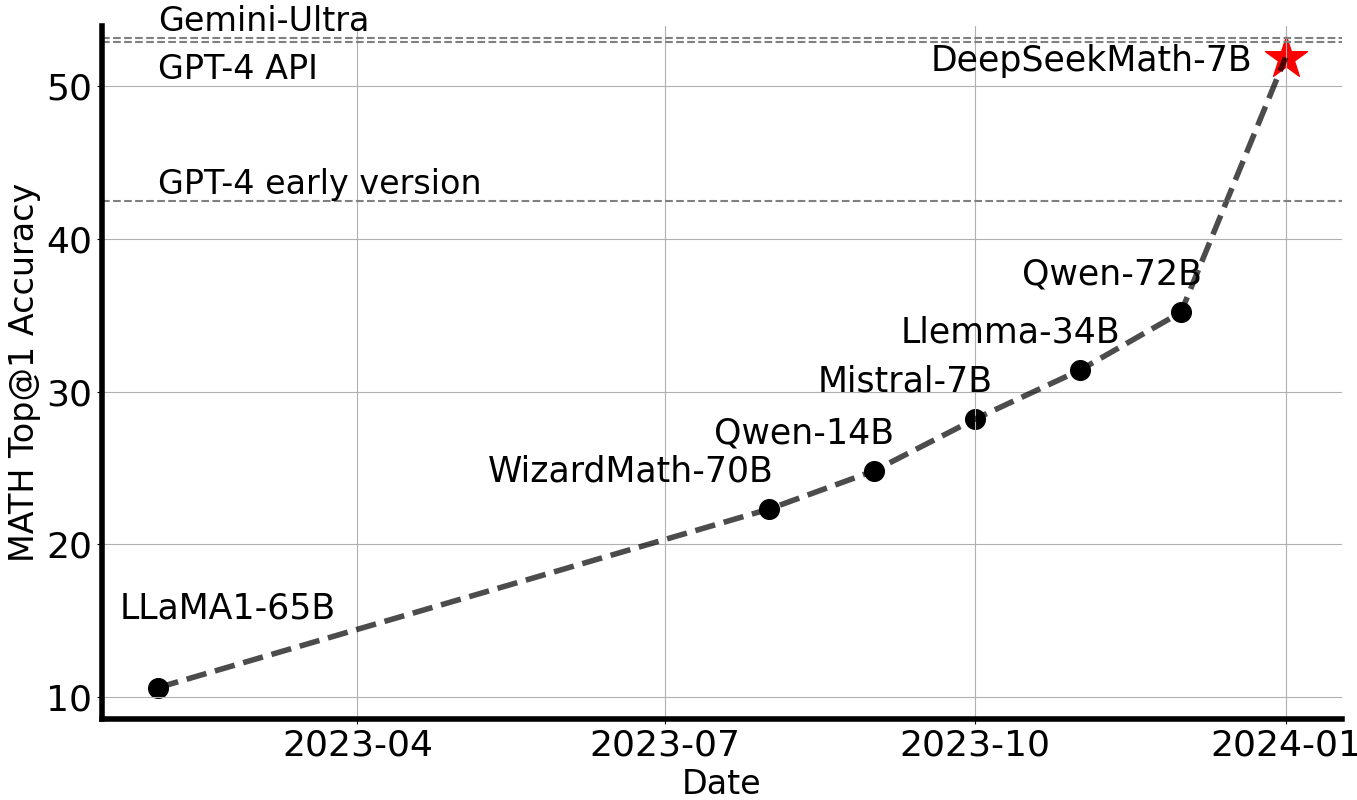}
 \caption{ Top\@1 accuracy of open-source models on the competition-level MATH benchmark \citep{MATH} without the use of external toolkits and voting techniques.
 % The Majority voting of \spmath~with 64 candidates achieves 61.9\% on MATH.
 }
\end{center}
\end{figure} \vspace{-3mm}
\newpage

% \begin{spacing}{0.9}
% \tableofcontents
% \end{spacing}

\newpage

\section{Introduction}

Large language models (LLM) have revolutionized the approach to mathematical reasoning in artificial intelligence, spurring significant advancements in both the quantitative reasoning benchmark \citep{MATH} and the geometry reasoning benchmark \citep{trinh2024solving}. Moreover, these models have proven instrumental in assisting humans in solving complex mathematical problems \citep{tao}. However, cutting-edge models such as GPT-4 \citep{gpt4} and Gemini-Ultra \citep{gemini} are not publicly available, and the currently accessible open-source models considerably trail behind in performance.

In this study, we introduce \spmath, a domain-specific language model that significantly outperforms the mathematical capabilities of open-source models and approaches the performance level of GPT-4 on academic benchmarks.
To achieve this, we create the DeepSeekMath Corpus, a large-scale high-quality pre-training corpus comprising 120B math tokens.
This dataset is extracted from the Common Crawl (CC) using a fastText-based classifier \citep{joulin2016fasttext}. In the initial iteration, the classifier is trained using instances from OpenWebMath \citep{openwebmath} as positive examples, while incorporating a diverse selection of other web pages to serve as negative examples. Subsequently, we employ the classifier to mine additional positive instances from the CC, which are further refined through human annotation. The classifier is then updated with this enhanced dataset to improve its performance. The evaluation results indicate that the large-scale corpus is of high quality, as our base model \spmath-Base 7B achieves 64.2\% on GSM8K \citep{gsm8k} and 36.2\% on the competition-level MATH dataset \citep{MATH}, outperforming Minerva 540B \citep{minerva}. In addition, the DeepSeekMath Corpus is multilingual, so we notice an improvement in Chinese mathematical benchmarks \citep{wei2023cmath,agieval}.
We believe that our experience in mathematical data processing is a starting point for the research community, and there is significant room for improvement in the future.

\spmath-Base is initialized with DeepSeek-Coder-Base-v1.5 7B \citep{deepseek-coder}, as we notice that starting from a code training model is a better choice compared to a general LLM. Furthermore, we observe the math training also improves model capability on MMLU \citep{mmlu} and BBH benchmarks \citep{bbh}, indicating it does not only enhance the model's mathematical abilities but also amplifies general reasoning capabilities.

After pre-training, we apply mathematical instruction tuning to \spmath-Base with chain-of-thought \citep{cot}, program-of-thought \citep{pot,pal}, and tool-integrated reasoning \citep{tora} data.
The resulting model \spmath-Instruct 7B beats all 7B counterparts and is comparable with 70B open-source instruction-tuned models.  

Furthermore, we introduce the Group Relative Policy Optimization (GRPO), a variant reinforcement learning (RL) algorithm of Proximal Policy Optimization (PPO) \citep{schulman2017proximal}.
GRPO foregoes the critic model, instead estimating
the baseline from group scores, significantly reducing training resources.
By solely using a subset of English instruction tuning data, GRPO obtains a substantial improvement over the strong DeepSeekMath-Instruct, including both in-domain (GSM8K: 82.9\% $\rightarrow$ 88.2\%, MATH: 46.8\% $\rightarrow$ 51.7\%) and out-of-domain mathematical tasks (e.g., CMATH: 84.6\% $\rightarrow$ 88.8\%) during the reinforcement learning phase.
We also provide a unified paradigm to understand different methods, such as Rejection Sampling Fine-Tuning (RFT) \citep{yuan2023scaling}, Direct Preference Optimization (DPO) \citep{dpo}, PPO and GRPO.
Based on such a unified paradigm, we find that all these methods are conceptualized as either direct or simplified RL techniques. 
We also conduct extensive experiments, e.g., online v.s. offline training, outcome v.s. process supervision, single-turn v.s. iterative RL and so on, to deeply investigate the essential elements of this paradigm.
At last, we explain why our RL boosts the performance of instruction-tuned models, and further summarize potential directions to achieve more effective RL based on this unified paradigm.

\subsection{Contributions}
Our contribution includes scalable math pre-training, along with the exploration and analysis of reinforcement learning.

\noindent
\textbf{Math Pre-Training at Scale}
\begin{itemize}[topsep=0pt]
    \item Our research provides compelling evidence that the publicly accessible Common Crawl data contains valuable information for mathematical purposes.
    By implementing a meticulously designed data selection pipeline, we successfully construct the DeepSeekMath Corpus, a high-quality dataset of 120B tokens from web pages filtered for mathematical content, which is almost 7 times the size of the math web pages used by Minerva \citep{minerva} and 9 times the size of the recently released OpenWebMath \citep{openwebmath}.

    \item Our pre-trained base model \spmath-Base 7B achieves comparable performance with Minerva 540B \citep{minerva}, indicating the number of parameters is not the only key factor in mathematical reasoning capability.
    A smaller model pre-trained on high-quality data could achieve strong performance as well. 

    \item We share our findings from math training experiments.
    Code training prior to math training improves models' ability to solve mathematical problems both with and without tool use.
    This offers a partial answer to the long-standing question: \textit{does code training improve reasoning abilities?}
    We believe it does, at least for mathematical reasoning.

    \item Although training on arXiv papers is common, especially in many math-related papers, it brings no notable improvements on all mathematical benchmarks adopted in this paper.
\end{itemize}

\noindent
\textbf{Exploration and Analysis of Reinforcement Learning}
\begin{itemize}[topsep=0pt]
    \item We introduce Group Relative Policy Optimization (GRPO), an efficient and effective reinforcement learning algorithm. GRPO foregoes the critic model, instead estimating the baseline from group scores, significantly reducing training resources compared to Proximal Policy Optimization (PPO).
    \item We demonstrate that GRPO significantly enhances the performance of our instruction-tuned model DeepSeekMath-Instruct, by solely using the instruction-tuning data. 
    Furthermore, we observe enhancements in the out-of-domain performance during the reinforcement learning process.
    \item We provide a unified paradigm to understand different methods, such as RFT, DPO, PPO, and GRPO. We also conduct extensive experiments, e.g., online v.s. offline training, outcome v.s. process supervision, single-turn v.s. iterative reinforcement learning, and so on to deeply investigate the essential elements of this paradigm.
    \item Based on our unified paradigm, we explore the reasons behind the effectiveness of reinforcement learning, and summarize several potential directions to achieve more effective reinforcement learning of LLMs.
\end{itemize}

\subsection{Summary of Evaluations and Metrics}
\begin{itemize}[topsep=0pt]
    \item \textbf{English and Chinese Mathematical Reasoning}:
    We conduct comprehensive assessments of our models on English and Chinese benchmarks, covering mathematical problems from grade-school level to college level.
    English benchmarks include GSM8K \citep{gsm8k}, MATH \citep{MATH}, SAT \citep{llemma}, OCW Courses \citep{minerva}, MMLU-STEM \citep{mmlu}.
    Chinese benchmarks include MGSM-zh \citep{mgsm}, CMATH \citep{wei2023cmath}, Gaokao-MathCloze \citep{agieval}, and Gaokao-MathQA \citep{agieval}.
    We evaluate models' ability to generate self-contained text solutions without tool use, and also the ability to solve problems using Python.
    
    On English benchmarks, \spmath-Base is competitive with the closed-source Minerva 540B \citep{minerva}, and surpasses all open-source base models (e.g., Mistral 7B \citep{mistral} and Llemma-34B \citep{llemma}), regardless of whether they've undergone math pre-training or not, often by a significant margin.
    Notably, \spmath-Base is superior on Chinese benchmarks, likely because we don't follow previous works \citep{minerva,llemma} to collect English-only math pre-training data, and also include high-quality non-English ones.
    With mathematical instruction tuning and reinforcement learning, the resulting \spmath-Instruct and \spmath-RL demonstrate strong performance, obtaining an accuracy of over 50\% on the competition-level MATH dataset for the first time within the open-source community.
    % competitive with the proprietary models GPT-4 and Gemini Ultra
    \item \textbf{Formal Mathematics}:
    We evaluate \spmath-Base using the informal-to-formal theorem proving task from \citep{dsp_proof} on miniF2F \citep{minif2f} with Isabelle \citep{isabelle} chosen to be the proof assistant.
    \spmath-Base demonstrates strong few-shot autoformalization performance.
    \item \textbf{Natural Language Understanding, Reasoning, and Code}:
    To build a comprehensive profile of models' general understanding, reasoning, and coding capabilities, we evaluate \spmath-Base on the Massive Multitask Language Understanding (MMLU) benchmark \citep{mmlu} which encompasses 57 multiple-choice tasks covering diverse subjects, BIG-Bench Hard (BBH) \citep{bbh} which consists of 23 challenging tasks that mostly require multi-step reasoning to solve, as well as HumanEval \citep{codex} and MBPP \citep{mbpp} which are widely used to evaluate code language models.
    Math pre-training benefits both language understanding and reasoning performance.
\end{itemize}

\section{Math Pre-Training}

\subsection{Data Collection and Decontamination}
\begin{figure}[h]
\begin{center}
        \includegraphics[width=0.99\textwidth]{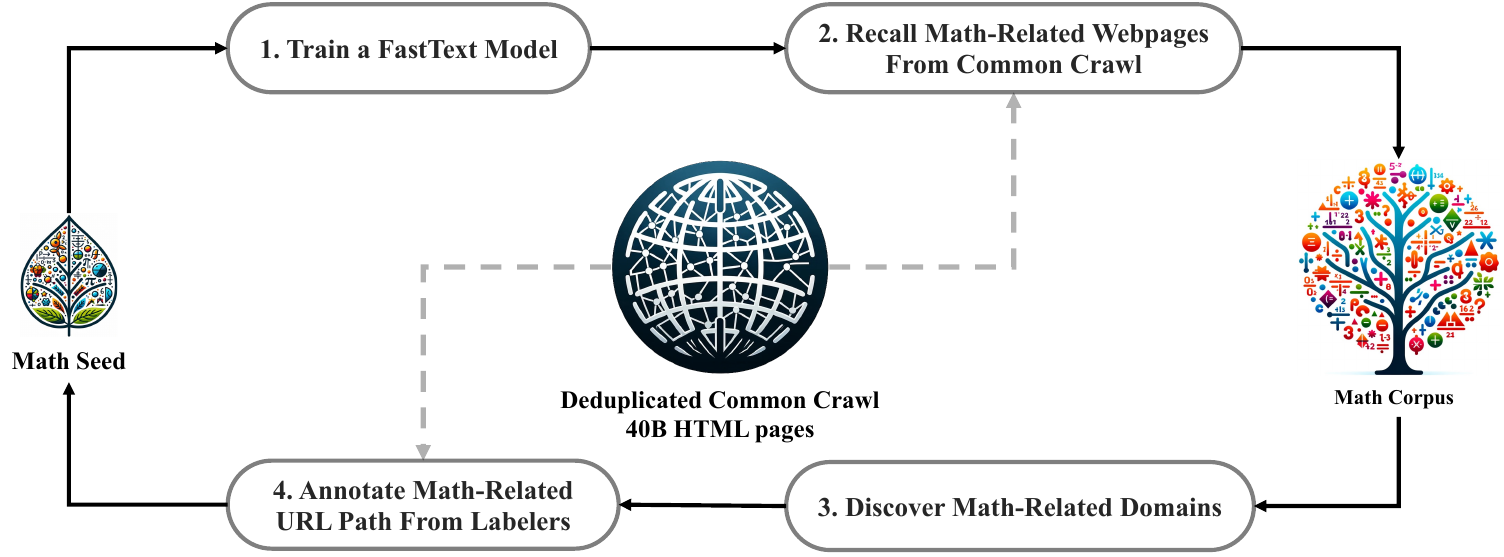}
 \caption{\centering An iterative pipeline that collects mathematical web pages from Common  Crawl. }
 \label{fig:data_collect}
\end{center}
\end{figure}

In this section, we will outline the process of constructing the DeepSeekMath Corpus from Common Crawl.
As depicted in Figure \ref{fig:data_collect}, we present an iterative pipeline that demonstrates how to systematically gather a large-scale mathematical corpus from Common Crawl, starting with a seed corpus (e.g., a small but high-quality collection of math-related dataset).
It's worth noting that this approach is also applicable to other domains, such as coding.

First, we choose OpenWebMath \citep{openwebmath}, a collection of high-quality mathematical web texts, as our initial seed corpus.
Using this corpus, we train a fastText model \citep{joulin2016fasttext} to recall more OpenWebMath-like mathematical web pages.
Specifically, we randomly select 500,000 data points from the seed corpus as positive training examples and another 500,000 web pages from Common Crawl as negative ones.
We employ an open-source library\footnote{\url{https://fasttext.cc}} for training, configuring the vector dimension to 256, learning rate to 0.1, the maximum length of word n-gram to 3, the minimum number of word occurrences to 3, and the number of training epochs to 3.
To reduce the size of the original Common Crawl, we employ URL-based deduplication and near-deduplication techniques, resulting in 40B HTML web pages.
We then recall mathematical web pages from deduplicated Common Crawl with the fastText model.
To filter out low-quality mathematical content, we rank the collected pages according to their scores predicted by the fastText model, and only preserve the top-ranking ones.
The volume of data preserved is assessed through pre-training experiments on the top 40B, 80B, 120B, and 160B tokens.
In the first iteration, we choose to keep the top 40B tokens.

After the first iteration of data collection, numerous mathematical web pages remain uncollected, mainly because the fastText model is trained on a set of positive examples that lacks sufficient diversity.
We therefore identify additional mathematical web sources to enrich the seed corpus, so that we can optimize the fastText model.
Specifically, we first organize the entire Common Crawl into disjoint domains;
a domain is defined as web pages sharing the same base URL.
For each domain, we calculate the percentage of web pages that are collected in the first iteration.
Domains where over 10\% of the web pages have been collected are classified as math-related (e.g., \url{mathoverflow.net}).
Subsequently, we manually annotate the URLs associated with mathematical content within these identified domains (e.g., \url{mathoverflow.net/questions}).
Web pages linked to these URLs, yet uncollected, will be added to the seed corpus.
This approach enables us to gather more positive examples, thereby training an improved fastText model capable of recalling more mathematical data in the subsequent iteration.
After four iterations of data collection, we end up with 35.5M mathematical web pages, totaling 120B tokens.
In the fourth iteration, we notice that nearly 98\% of the data has already been collected in the third iteration, so we decide to cease data collection.

To avoid benchmark contamination, we follow \cite{deepseek-coder} to filter out web pages containing questions or answers from English mathematical benchmarks such as GSM8K \citep{gsm8k} and MATH \citep{MATH} and Chinese benchmarks such as CMATH \citep{wei2023cmath} and AGIEval \citep{agieval}.
The filtering criteria are as follows: any text segment containing a 10-gram string that matches exactly with any sub-string from the evaluation benchmarks is removed from our math training corpus.
For benchmark texts that are shorter than 10 grams but have at least 3 grams, we employ exact matching to filter out contaminated web pages.

\subsection{Validating the Quality of the \spmath~Corpus}

We run pre-training experiments to investigate how the \spmath~Corpus is compared with the recently released math-training corpora:
\begin{itemize}[topsep=0pt]
    \item \textbf{MathPile} \citep{mathpile}: a multi-source corpus (8.9B tokens) aggregated from textbooks, Wikipedia, ProofWiki, CommonCrawl, StackExchange, and arXiv, with the majority (over 85\%) sourced from arXiv;
    \item \textbf{OpenWebMath} \citep{openwebmath}: CommonCrawl data filtered for mathematical content, totaling 13.6B tokens;
    \item \textbf{Proof-Pile-2} \citep{llemma}: a mathematical corpus consisting of OpenWebMath, AlgebraicStack (10.3B tokens of mathematical code), and arXiv papers (28.0B tokens).
    When experimenting on Proof-Pile-2, we follow \cite{llemma} to use an arXiv:Web:Code ratio of 2:4:1.
\end{itemize}

\subsubsection{Training Setting}
\label{sec:quality-policy}
We apply math training to a general pre-trained language model with 1.3B parameters, which shares the same framework as the DeepSeek LLMs \citep{deepseek-llm}, denoted as DeepSeek-LLM 1.3B.
We separately train a model on each mathematical corpus for 150B tokens. All experiments are conducted using the efficient and light-weight HAI-LLM \citep{haillm} training framework.
Following the training practice of DeepSeek LLMs, we use the AdamW optimizer \citep{adamW} with $\beta_1=0.9$, $\beta_2=0.95$, and $\mathrm{weight\_decay}=0.1$, along with a multi-step learning rate schedule where the learning rate reaches the peak after 2,000 warmup steps, decreases to its 31.6\% after 80\% of the training process, and further decreases to 10.0\% of the peak after 90\% of the training process.
We set the maximum value of learning rate to 5.3e-4, and use a batch size of 4M tokens with a 4K context length.

\begingroup
\setlength{\tabcolsep}{3pt} % Default value: 6pt
\renewcommand{\arraystretch}{1} % Default value: 1
\begin{table*}[h]
    \centering
    \adjustbox{max width=\textwidth}{
\begin{tabular}{llcccccccc} 
\toprule
\multirow{3}{*}{Math Corpus} & \multirow{3}{*}{Size} & \multicolumn{5}{c}{English Benchmarks}                        & \multicolumn{3}{c}{Chinese Benchmarks}                \\ 
\cmidrule(lr){3-7}\cmidrule(lr){8-10}
                             & & GSM8K  & MATH   & OCW & SAT    & \begin{tabular}[c]{@{}c@{}}MMLU \\ STEM\end{tabular} & CMATH  & \begin{tabular}[c]{@{}c@{}}Gaokao \\ MathCloze\end{tabular} & \begin{tabular}[c]{@{}c@{}}Gaokao \\ MathQA\end{tabular}  \\ \midrule
No Math Training & N/A             & 2.9\%  & 3.0\%  & 2.9\%       & 15.6\% & 19.5\%    & 12.3\% & 0.8\%            & 17.9\%         \\ 
\midrule
MathPile & 8.9B                     & 2.7\%  & 3.3\%  & 2.2\%       & 12.5\% & 15.7\%    & 1.2\%  & 0.0\%            & 2.8\%          \\
OpenWebMath & 13.6B                  & 11.5\% & 8.9\%  & 3.7\%       & 31.3\% & 29.6\%    & 16.8\% & 0.0\%            & 14.2\%         \\
Proof-Pile-2 & 51.9B                 & 14.3\% & 11.2\% & 3.7\%       & 43.8\% & 29.2\%    & 19.9\% & 5.1\%            & 11.7\%         \\ 
\midrule
\spmath~Corpus & \textbf{120.2B}          & \textbf{23.8\%} & \textbf{13.6\%} & \textbf{4.8\%}       & \textbf{56.3\%} & \textbf{33.1\%}    & \textbf{41.5\%} & \textbf{5.9\%}            & \textbf{23.6\%}         \\
\bottomrule
\end{tabular}
    }
    \caption{
    Performance of DeepSeek-LLM 1.3B trained on different mathematical corpora, evaluated using few-shot chain-of-thought prompting.
    Corpus sizes are calculated using our tokenizer with a vocabulary size of 100K.
    }
    \label{tab:corpora_comparison}
\end{table*}
\endgroup

\begin{figure*}[t!]
    \centering
    \includegraphics[width=0.88\textwidth]{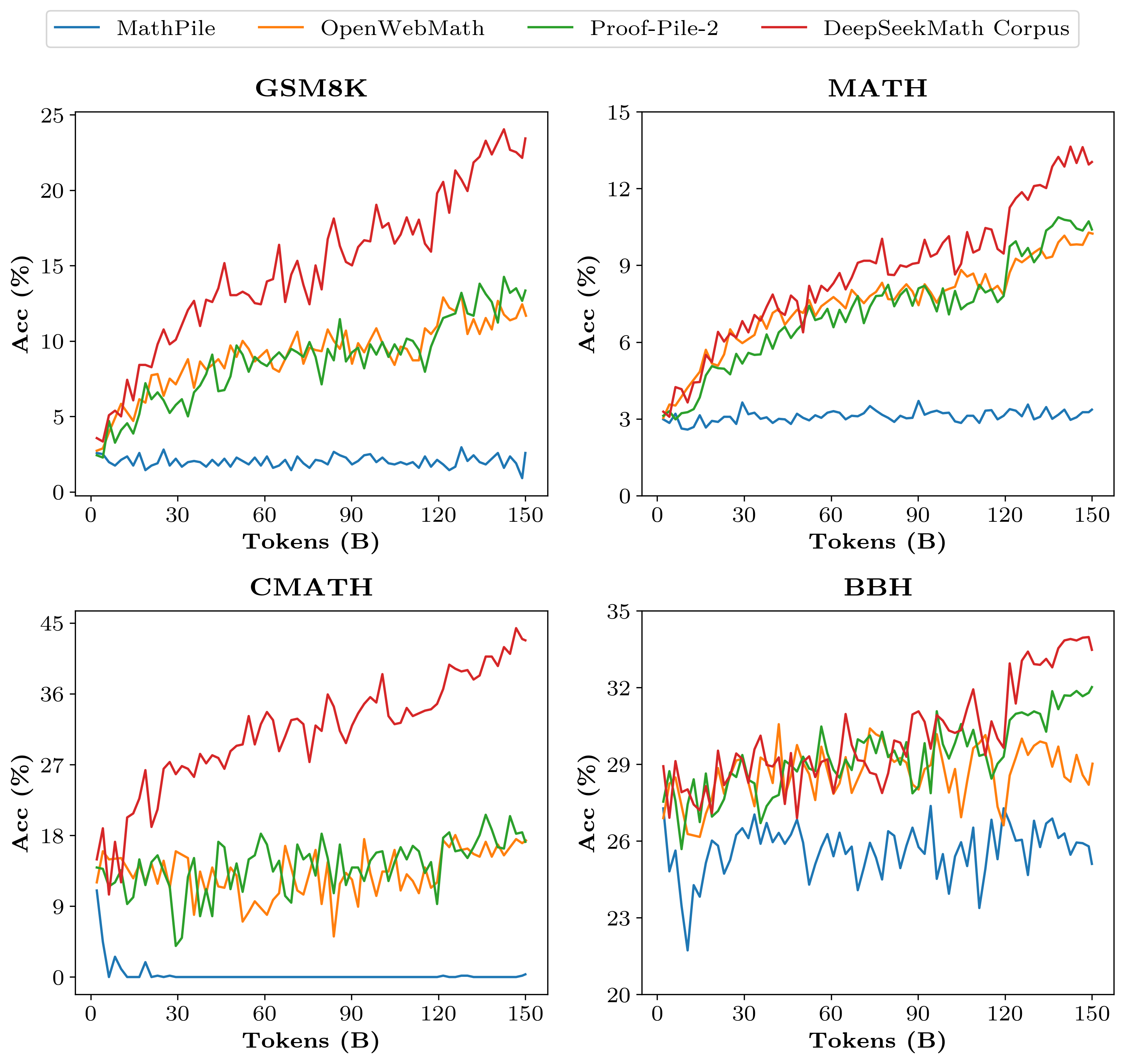}
    \caption{\centering Benchmark curves of DeepSeek-LLM 1.3B trained on different mathematical corpora.}
    \label{fig:corpora_comparisons}
\end{figure*}
\subsubsection{Evaluation Results}

\textbf{The \spmath~Corpus is of high quality, covers multilingual mathematical content, and is the largest in size.}
\begin{itemize}[topsep=0pt]
    \item \textbf{High-quality}:
    We evaluate downstream performance on 8 mathematical benchmarks using few-shot chain-of-thought prompting \cite{cot}.
    As shown in Table \ref{tab:corpora_comparison}, there is a clear performance lead of the model trained on the \spmath~Corpus. Figure \ref{fig:corpora_comparisons} shows that the model trained on the DeepSeekMath Corpus demonstrates better performance than Proof-Pile-2 at 50B tokens (1 full epoch of Proof-Pile-2), indicating the average quality of DeepSeekMath Corpus is higher. 
    \item \textbf{Multilingual}:
    The \spmath~Corpus encompasses data in multiple languages, predominantly featuring English and Chinese as the two most represented languages.
    As shown in Table \ref{tab:corpora_comparison}, training on the \spmath~Corpus enhances mathematical reasoning performance in both English and Chinese.
    In contrast, existing mathematical corpora, which are primarily English-centric, show limited improvement and may even hinder performance in Chinese mathematical reasoning.
    \item \textbf{Large-scale}:
    The \spmath~Corpus is several times larger than existing mathematical corpora.
    As depicted in Figure \ref{fig:corpora_comparisons}, DeepSeek-LLM 1.3B, when trained on the \spmath~Corpus, shows a steeper learning curve along with more lasting improvements.
    In contrast, the baseline corpora are much smaller, and have already been repeated multiple rounds during training, with the resulting model performance quickly reaching a plateau.
\end{itemize}

\subsection{Training and Evaluating \spmath-Base 7B}

In this section, we introduce \spmath-Base 7B, a base model with strong reasoning abilities, especially in mathematics.
Our model is initialized with DeepSeek-Coder-Base-v1.5 7B \citep{deepseek-coder} and trained for 500B tokens. The distribution of the data  is as follows: 56\% is from the DeepSeekMath Corpus, 4\% from AlgebraicStack, 10\% from arXiv, 20\% is Github code, and the remaining 10\% is natural language data from Common Crawl in both English and Chinese.
We mainly adopt the training setting specified in Section \ref{sec:quality-policy}, except that we set the maximum value of the learning rate to 4.2e-4 and use a batch size of 10M tokens.

We conduct a comprehensive assessment of the mathematical capabilities of \spmath-Base 7B, focusing on its ability to produce self-contained mathematical solutions without relying on external tools, solve mathematical problems using tools, and conduct formal theorem proving.
Beyond mathematics, we also provide a more general profile of the base model, including its performance of natural language understanding, reasoning, and programming skills.

\paragraph{Mathematical Problem Solving with Step-by-Step Reasoning}

We evaluate \spmath-Base's performance of solving mathematical problems using few-shot chain-of-thought prompting \citep{cot}, across eight benchmarks in English and Chinese.
These benchmarks encompass quantitative reasoning (e.g., GSM8K \citep{gsm8k}, MATH \citep{MATH}, and CMATH \citep{wei2023cmath}) and multiple-choice problems (e.g., MMLU-STEM \citep{mmlu} and Gaokao-MathQA \citep{agieval}), covering diverse fields of mathematics from elementary to college-level complexity.

As shown in Table \ref{tab:base_math_cot}, \spmath-Base 7B leads in performance across all eight benchmarks among the open-source base models (including the widely-used general model Mistral 7B \citep{mistral} and the recently released Llemma 34B \citep{llemma} which underwent math training on Proof-Pile-2 \citep{llemma}).
Notably, on the competition-level MATH dataset, \spmath-Base surpasses existing open-source base models by over 10\% absolute, and outperforms Minerva 540B \citep{minerva}, a closed-source base model 77 times larger which builds on PaLM \citep{palm} and is further trained on mathematical texts.

\begingroup
\setlength{\tabcolsep}{3pt} % Default value: 6pt
\renewcommand{\arraystretch}{1} % Default value: 1
\begin{table*}[h]
    \centering
    \adjustbox{max width=\textwidth}{
\begin{tabular}{llcccccccc} 
\toprule
\multicolumn{1}{l}{\multirow{3}{*}{Model}} &\multicolumn{1}{l}{\multirow{3}{*}{Size}}& \multicolumn{5}{c}{English Benchmarks}                                                             & \multicolumn{3}{c}{Chinese Benchmarks}     \\\cmidrule(lr){3-7}\cmidrule(lr){8-10}
& & GSM8K           & MATH            & OCW      & SAT             & \makecell{MMLU \\ STEM}       & CMATH  & \makecell{Gaokao \\ MathCloze} & \makecell{Gaokao \\ MathQA} \\ 
\midrule

\multicolumn{10}{c}{Closed-Source Base Model} \\
\midrule
Minerva & 7B          & 16.2\%          & 14.1\%          & 7.7\%           & -               & 35.6\%          & -               & -                & -                \\
Minerva& 62B                  & 52.4\%          & 27.6\%          & 12.0\%          & -               & 53.9\%          & -               & -                & -                \\
Minerva& 540B                & 58.8\%          & 33.6\%          & 17.6\%          & -               & 63.9\%          & -               & -                & -                \\ 
\midrule

\multicolumn{10}{c}{Open-Source Base Model} \\
\midrule
Mistral & 7B                             & 40.3\%          & 14.3\%          & 9.2\%           & 71.9\%          & 51.1\%          & 44.9\%          & 5.1\%            & 23.4\%           \\
\midrule
Llemma & 7B & 37.4\%          & 18.1\%          & 6.3\%           & 59.4\%          & 43.1\%~         & 43.4\%          & 11.9\%           & 23.6\%           \\
Llemma& 34B                 & 54.0\%          & 25.3\%          & 10.3\%          & 71.9\%          & 52.9\%          & 56.1\%          & 11.9\%           & 26.2\%           \\ 
\midrule
\spmath-Base & 7B           & \textbf{64.2\%} & \textbf{36.2\%} & \textbf{15.4\%} & \textbf{84.4\%} & \textbf{56.5\%} & \textbf{71.7\%} & \textbf{20.3\%}  & \textbf{35.3\%}  \\
\bottomrule
\end{tabular}
}
    \caption{
    Comparisons between \spmath-Base 7B and strong base models on English and Chinese mathematical benchmarks.
    Models are evaluated with chain-of-thought prompting.
    Minerva results are quoted from \cite{minerva}.
    }
    \label{tab:base_math_cot}
\end{table*}
\endgroup

\paragraph{Mathematical Problem Solving with Tool Use}
We evaluate program-aided mathematical reasoning on GSM8K and MATH using few-shot program-of-thought prompting \citep{pot,pal}.
Models are prompted to solve each problem by writing a Python program where libraries such as \textit{math} and \textit{sympy} can be utilized for intricate computations.
The execution result of the program is evaluated as the answer.
As shown in Table \ref{tab:base_math_pot_proof}, \spmath-Base 7B outperforms the prior state-of-the-art Llemma 34B.

\begingroup
\setlength{\tabcolsep}{3pt} % Default value: 6pt
\renewcommand{\arraystretch}{1} % Default value: 1
\begin{table*}[htb]
    \centering
\begin{small}
\begin{tabular}{llcccc} 
\toprule
\multicolumn{1}{l}{\multirow{2}{*}{Model}}  & \multicolumn{1}{l}{\multirow{2}{*}{Size}} &\multicolumn{2}{c}{Problem Solving w/ Tools}                                                             & \multicolumn{2}{c}{Informal-to-Formal Proving}           \\ \cmidrule(lr){3-4}\cmidrule(lr){5-6}
& & GSM8K+Python           & MATH+Python     & miniF2F-valid    & miniF2F-test    \\ 
\midrule
Mistral & 7B                           & 48.5\%          & 18.2\%       &      18.9\%         &  18.0\%          \\ 
\midrule
CodeLlama & 7B    & 27.1\%          & 17.2\%       & 16.3\%        & 17.6\%         \\
CodeLlama& 34B                                 & 52.7\%          & 23.5\%      & 18.5\%        & 18.0\%       \\ 
\midrule
Llemma & 7B  & 41.0\%          & 18.6\%     & 20.6\%        & 22.1\%        \\
Llemma& 34B                                 & 64.6\% & 26.3\%      & 21.0\%        & 21.3\%         \\ 
\midrule
\spmath-Base & 7B                & \textbf{66.9\%}          & \textbf{31.4\%}  & \textbf{25.8\%}        & \textbf{24.6\%}    \\
\bottomrule
\end{tabular}
\end{small}
    \caption{
    Few-shot evaluation of base models' ability to solve mathematical problems using tools and the ability to conduct informal-to-formal theorem proving in Isabelle.
    }
    \label{tab:base_math_pot_proof}
\end{table*}
\endgroup

\paragraph{Formal Mathematics}

Formal proof automation is beneficial to ensure the accuracy and reliability of mathematical proofs and enhance efficiency, with increasing attention in recent years.
We evaluate \spmath-Base 7B on the task of informal-to-formal proving from \citep{dsp_proof} which is to generate a formal proof based on an informal statement, a formal counterpart of the statement, and an informal proof.
We evaluate on miniF2F \citep{minif2f}, a benchmark for formal Olympiad-level mathematics, and generate a formal proof in Isabelle for each problem with few-shot prompting.
Following \cite{dsp_proof}, we leverage models to generate proof sketches, and execute the off-the-shelf automated prover Sledgehammer \citep{sledgehammer} to fill in the missing details.
As shown in Table \ref{tab:base_math_pot_proof}, \spmath-Base 7B demonstrates strong performance in proof autoformalization.

\begingroup
\setlength{\tabcolsep}{3pt} % Default value: 6pt
\renewcommand{\arraystretch}{1} % Default value: 1
\begin{table*}[htb]
    \centering
\begin{small}

\begin{tabular}{llcccc} 
\toprule
\multicolumn{1}{l}{Model} &\multicolumn{1}{l}{Size}                      & MMLU   & BBH    & HumanEval (Pass@1) & MBPP (Pass@1)  \\ 
\midrule
Mistral & 7B                      & \textbf{62.4\%} & 55.7\% & 28.0\%             & 41.4\%         \\
\midrule
DeepSeek-Coder-Base-v1.5\textsuperscript{$\dag$} & 7B & 42.9\% & 42.9\% & 40.2\%             & 52.6\%         \\
DeepSeek-Coder-Base-v1.5 & 7B          & 49.1\% & 55.2\% & \textbf{43.2\%}             & \textbf{60.4\%}         \\ 
\midrule
\spmath-Base & 7B            & 54.9\% & \textbf{59.5\%} & 40.9\%             & 52.6\%         \\
\bottomrule
\end{tabular}
\end{small}
    \caption{
    Evaluation on natural language understanding, reasoning, and code benchmarks.
    DeepSeek-Coder-Base-v1.5\textsuperscript{$\dag$} is the checkpoint right before learning rate decay, which is used to train \spmath-Base.
    On MMLU and BBH, we use few-shot chain-of-thought prompting.
    On HumanEval and MBPP, we evaluate model performance under the zero-shot setting and a few-shot setting, respectively.
    }
    \label{tab:understanding_reasoning_code}
\end{table*}
\endgroup

\paragraph{Natural Language Understanding, Reasoning, and Code}

We evaluate model performance of natural language understanding on MMLU \citep{mmlu}, reasoning on BBH \citep{bbh}, and coding capabilities on HumanEval \citep{codex} and MBPP \citep{mbpp}. As shown in Table \ref{tab:understanding_reasoning_code}, \spmath-Base 7B exhibits significant enhancements in performance on MMLU and BBH over its precursor, DeepSeek-Coder-Base-v1.5 \citep{deepseek-coder}, illustrating the positive impact of math training on language understanding and reasoning.
Additionally, by including code tokens for continual training, \spmath-Base 7B effectively maintains the performance of DeepSeek-Coder-Base-v1.5 on the two coding benchmarks.
Overall, \spmath-Base 7B significantly outperforms the general model Mistral 7B \citep{mistral} on the three reasoning and coding benchmarks.

\section{Supervised Fine-Tuning}

\subsection{SFT Data Curation}

We construct a mathematical instruction-tuning dataset covering English and Chinese problems from different mathematical fields and of varying complexity levels:
problems are paired with solutions in chain-of-thought (CoT) \citep{cot}, program-of-thought (PoT) \citep{pot,pal}, and tool-integrated reasoning format \citep{tora}.
The total number of training examples is 776K.
\begin{itemize}[topsep=0pt]
    \item \textbf{English mathematical datasets}:
    We annotate GSM8K and MATH problems with tool-integrated solutions, and adopt a subset of MathInstruct \citep{MathInstruct} along with the training set of Lila-OOD \citep{lila} where problems are solved with CoT or PoT.
    Our English collection covers diverse fields of mathematics, e.g., algebra, probability, number theory, calculus, and geometry.
    \item \textbf{Chinese mathematical datasets}:
    We collect Chinese K-12 mathematical problems spanning 76 sub-topics such as linear equations, with solutions annotated in both CoT and tool-integrated reasoning format.
\end{itemize}

\subsection{Training and Evaluating \spmath-Instruct 7B}

In this section, we introduce \spmath-Instruct 7B which undergoes mathematical instruction tuning based on \spmath-Base.
Training examples are randomly concatenated until reaching a maximum context length of 4K tokens.
We train the model for 500 steps with a batch size of 256 and a constant learning rate of 5e-5.

We evaluate models' mathematical performance both without and with tool use, on 4 quantitative reasoning benchmarks in English and Chinese.
We benchmark our model against the leading models of the time:
\begin{itemize}[topsep=0pt]
    \item \textbf{Closed-source models} include:
    (1) the GPT family among which GPT-4 \citep{gpt4} and GPT-4 Code Interpreter~\footnote{\url{https://openai.com/blog/chatgpt-plugins\#\#code-interpreter}} are the most capable ones,
    (2) Gemini Ultra and Pro \citep{gemini},
    (3) Inflection-2 \citep{inflection-2},
    (4) Grok-1~\footnote{\url{https://x.ai/model-card}},
    as well as models recently released by Chinese companies including
    (5) Baichuan-3~\footnote{\url{https://www.baichuan-ai.com}},
    (6) the latest GLM-4~\footnote{\url{https://open.bigmodel.cn/dev/api\#glm-4}}
    % and ChatGLM-Turbo~\footnote{\url{https://open.bigmodel.cn/dev/api\#glm-3-turbo}}
    from the GLM family \citep{glm}.
    % and (7) Ernie-bot-4.0~\footnote{\url{https://yiyan.baidu.com/}}.
    These models are for general purposes, most of which have undergone a series of alignment procedures.
    \item \textbf{Open-source models} include:
    general models like (1) DeepSeek-LLM-Chat 67B \citep{deepseek-llm}, (2) Qwen 72B \citep{qwen}, (3) SeaLLM-v2 7B \citep{seallm}, and (4) ChatGLM3 6B \citep{chatglm3},
    as well as models with enhancements in mathematics including
    (5) InternLM2-Math 20B~\footnote{\url{https://github.com/InternLM/InternLM-Math}} which builds on InternLM2 and underwent math training followed by instruction tuning,
    (6) Math-Shepherd-Mistral 7B which applys PPO training \citep{schulman2017proximal} to Mistral 7B \citep{mistral} with a process-supervised reward model,
    (7) the WizardMath series \citep{wizardmath} which improves mathematical reasoning in Mistral 7B and Llama-2 70B \citep{llama2} using evolve-instruct (i.e., a version of instruction tuning that uses AI-evolved instructions) and PPO training with training problems primarily sourced from GSM8K and MATH,
    (8) MetaMath 70B \citep{metamath} which is Llama-2 70B fine-tuned on an augmented version of GSM8K and MATH,
    (9) ToRA 34B \cite{tora} which is CodeLlama 34B fine-tuned to do tool-integrated mathematical reasoning,
    (10) MAmmoTH 70B \citep{MathInstruct} which is Llama-2 70B instruction-tuned on MathInstruct.
\end{itemize}

\begingroup
\setlength{\tabcolsep}{3pt} % Default value: 6pt
\renewcommand{\arraystretch}{1} % Default value: 1
\begin{table*}[t!]
    \centering
\adjustbox{max width=\textwidth}{
\begin{tabular}{llcccc} 
\toprule
\multicolumn{1}{l}{\multirow{2}{*}{Model}}  & \multicolumn{1}{l}{\multirow{2}{*}{Size}}                & \multicolumn{2}{c}{English Benchmarks}           & \multicolumn{2}{c}{Chinese Benchmarks}                                                        \\ 
\cmidrule(r){3-4} \cmidrule(r){5-6} 
& & GSM8K                          & MATH            & MGSM-zh         & CMATH   \\
\midrule
\midrule
\multicolumn{6}{c}{\textbf{Chain-of-Thought Reasoning}}                                                                                                                         \\ 
\midrule
\multicolumn{6}{c}{Closed-Source Model}                                                                                                                                         \\ 
\midrule
Gemini Ultra  &     -           & \textcolor{gray}{94.4\%} & 53.2\%               &        -              &        -                                                                    \\
GPT-4 &       -                & 92.0\%                         & 52.9\%               &         -             &   86.0\%                                                                \\
Inflection-2 &     -            & 81.4\%                         & 34.8\%               &            -          &       -                                                               \\
GPT-3.5 &    -                & 80.8\%                         & 34.1\%               &           -           &     73.8\%                                                         \\
Gemini Pro &     -             & \textcolor{gray}{86.5\%} & 32.6\%               &              -        &      -                                                                   \\
Grok-1 &          -            & 62.9\%                         & 23.9\%               &           -           &       -                                                                 \\
\midrule
Baichuan-3 &      -            & 88.2\%                         & 49.2\%               &            -          &         -                                                            \\
GLM-4 &           -            & 87.6\%                         & 47.9\%               &            -          &       -                                                          \\
\midrule
\multicolumn{6}{c}{Open-Source Model}                                                                                                                                           \\ 
\midrule
InternLM2-Math & 20B           & 82.6\%                         & 37.7\%          &           -      &             -                                                     \\
Qwen & 72B                     & 78.9\%                         & 35.2\%          &          -       &       -                                                         \\
Math-Shepherd-Mistral & 7B & 84.1\%                         & 33.0\%          &             -    &               -                                                     \\
WizardMath-v1.1 & 7B           & 83.2\%                         & 33.0\%          &            -     &               -                                                  \\
DeepSeek-LLM-Chat & 67B        & 84.1\%                         & 32.6\%               & 74.0\%               & 80.3\%                                                            \\
MetaMath & 70B                 & 82.3\%                         & 26.6\%          & 66.4\%          & 70.9\%                                                             \\
SeaLLM-v2 & 7B & 78.2\% & 27.5\% & 64.8\% & - \\
ChatGLM3 & 6B                  & 72.3\%                         & 25.7\%          &         -        &          -                                                      \\
WizardMath-v1.0 & 70B          & 81.6\%                         & 22.7\%          & 64.8\%          & 65.4\%                                                            \\ 
\midrule
\textbf{\spmath-Instruct} & 7B     & 82.9\%                         & 46.8\%          & 73.2\%          & 84.6\%                                                     \\
\textbf{\spmath-RL} & 7B           & \textbf{88.2\%}                & \textbf{51.7\%} & \textbf{79.6\%} & \textbf{88.8\%}                                        \\
\midrule
\midrule
\multicolumn{6}{c}{\textbf{Tool-Integrated Reasoning}}                                                                                                                          \\ 
\midrule
\multicolumn{6}{c}{Closed-Source Model}                                                                                                                                         \\ 
\midrule
GPT-4 Code Interpreter &   -   & 97.0\%                         & 69.7\%          &            -     &          -                                                      \\ 
\midrule
\multicolumn{6}{c}{Open-Source Model}                                                                                                                                           \\ 
\midrule
InternLM2-Math & 20B           & 80.7\%                         & 54.3\%          &          -       &      -                                                               \\
DeepSeek-LLM-Chat & 67B        & 86.7\%                & 51.1\%          & 76.4\%          & 85.4\%                                                            \\
ToRA & 34B                     & 80.7\%                         & 50.8\%          & 41.2\%          & 53.4\%                                                              \\
MAmmoTH & 70B                  & 76.9\%                         & 41.8\%          &          -       &            -                                                         \\ 
\midrule
\textbf{\spmath-Instruct} & 7B     & 83.7\%                         & 57.4\%          & 72.0\%          & 84.3\%                                                                \\
\textbf{\spmath-RL} & 7B           & \textbf{86.7\%}                         & \textbf{58.8\%} & \textbf{78.4\%} & \textbf{87.6\%}                                                       \\
\bottomrule
\end{tabular}
}
    \caption{
   Performance of Open- and Closed-Source models with both Chain-of-Thought and Tool-Integrated Reasoning on English and Chinese Benchmarks.
   Scores in \textcolor{gray}{gray} denote majority votes with 32 candidates; The others are Top1 scores.
   \spmath-RL 7B beats all open-source models from 7B to 70B, as well as the majority of closed-source models. Although \spmath-RL 7B is only further trained on chain-of-thought-format instruction tuning data of GSM8K and MATH, it improves over \spmath-Instruct 7B on all benchmarks.
   }
    \label{tab:sft_rl_math}
\end{table*}
\endgroup

As shown in Table \ref{tab:sft_rl_math}, under the evaluation setting where tool use is disallowed, \spmath-Instruct 7B demonstrates strong performance of step-by-step reasoning.
Notably, on the competition-level MATH dataset, our model surpasses all open-source models and the majority of proprietary models (e.g., Inflection-2 and Gemini Pro) by at least 9\% absolute.
This is true even for models that are substantially larger (e.g., Qwen 72B) or have been specifically enhanced through math-focused reinforcement learning (e.g., WizardMath-v1.1 7B).
While \spmath-Instruct rivals the Chinese proprietary models GLM-4 and Baichuan-3 on MATH, it still underperforms GPT-4 and Gemini Ultra.

Under the evaluation setting where models are allowed to integrate natural language reasoning and program-based tool use for problem solving, \spmath-Instruct 7B approaches an accuracy of 60\% on MATH, surpassing all existing open-source models.
On the other benchmarks, our model is competitive with DeepSeek-LLM-Chat 67B, the prior state-of-the-art that is 10 times larger.
\section{Reinforcement Learning}

\subsection{Group Relative Policy Optimization}
Reinforcement learning (RL) has been proven to be effective in further improving the mathematical reasoning ability of LLMs after the Supervised Fine-Tuning (SFT) stage \citep{wang2023math,wizardmath}.
In this section, we introduce our efficient and effective RL algorithm, Group Relative Policy Optimization (GRPO).

\subsubsection{From PPO to GRPO} Proximal Policy Optimization (PPO) \citep{schulman2017proximal} is an actor-critic RL algorithm that is widely used in the RL fine-tuning stage of LLMs \citep{ouyang2022training}. In particular, it optimizes LLMs by maximizing the following surrogate objective:
\begin{equation}
\footnotesize
    \mathcal{J}_{PPO}(\theta) = \mathbb{E}{[q \sim P(Q), o \sim \pi_{\theta_{old}}(O|q)]} \frac{1}{|o|} \sum_{t=1}^{|o|} \min \left[ \frac{\pi_\theta(o_{t} | q, o_{<t})}{\pi_{\theta_{old}}(o_{t} | q, o_{<t})} A_{t}, \text{clip} \left( \frac{\pi_\theta(o_{t} | q, o_{<t})}{\pi_{\theta_{old}}(o_{t} | q, o_{<t})}, 1 - \epsilon, 1 + \epsilon \right)  A_{t} \right] ,
\end{equation}
where $\pi_{\theta}$ and $\pi_{\theta_{old}}$ are the current and old policy models, and $q, o$  are questions and outputs sampled from the question dataset and the old policy  $\pi_{\theta_{old}}$, respectively.  $\epsilon$ is a clipping-related hyper-parameter introduced in PPO for stabilizing training.  $A_t$  is the advantage, which is computed by applying Generalized Advantage Estimation (GAE) \citep{gae}, based on the rewards  $\{r_{\ge t}\}$  and a learned value function  $V_{\psi}$. Thus, in PPO, a value function needs to be trained alongside the policy model and to mitigate over-optimization of the reward model, the standard approach is to add a per-token KL penalty from a reference model in the reward at each token \citep{ouyang2022training}, i.e., 
\begin{equation}
    r_{t} = r_\phi(q, o_{\le t}) - \beta \log\frac{\pi_{\theta}(o_{t}|q, o_{<t})}{\pi_{ref}(o_{t}|q, o_{<t})},
\label{eq:PPO-reward}
\end{equation}
where $r_\phi$  is the reward model, $\pi_{ref}$ is the reference model, which is usually the initial SFT model, and $\beta$  is the coefficient of the KL penalty.

\begin{figure*}[t]
\centering
\includegraphics[width=0.95\linewidth]{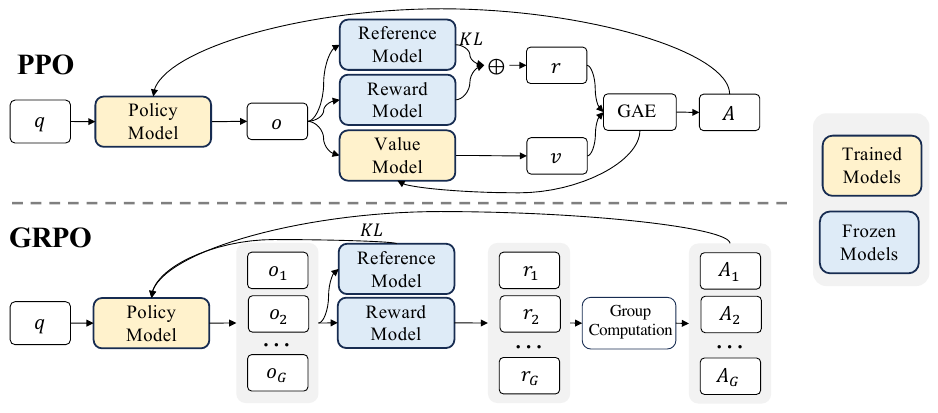}\vspace{-0.1in}
\caption{Demonstration of PPO and our GRPO. GRPO foregoes the value model, instead estimating the baseline from group scores, significantly reducing training resources.}
\label{fig:grpo}
\end{figure*}

As the value function employed in PPO is typically another model of comparable size as the policy model, it brings a substantial memory and computational burden. Additionally, during RL training, the value function is treated as a baseline in the calculation of the advantage for variance reduction. While in the LLM context, usually only the last token is assigned a reward score by the reward model, which may complicate the training of a value function that is accurate at each token. To address this, as shown in Figure \ref{fig:grpo}, we propose Group Relative Policy Optimization (GRPO), which obviates the need for additional value function approximation as in PPO, and instead uses the average reward of multiple sampled outputs, produced in response to the same question, as the baseline. More specifically, for each question $q$, GRPO samples a group of outputs $\{o_1, o_2, \cdots, o_G\}$  from the old policy  $\pi_{\theta_{old}}$  and then optimizes the policy model by maximizing the following objective:
\begin{equation}
\footnotesize
\begin{split}
    \mathcal{J}_{GRPO}(\theta) &= \mathbb{E}{[q \sim P(Q), \{o_i\}_{i=1}^G \sim \pi_{\theta_{old}}(O|q)]}  \\
    & \frac{1}{G}\sum_{i=1}^G\frac{1}{|o_i|} \sum_{t=1}^{|o_i|} \left\{ \min \left[ \frac{\pi_\theta(o_{i,t} | q, o_{i,<t})}{\pi_{\theta_{old}}(o_{i,t} | q, o_{i,<t})} \hat{A}_{i,t}, \text{clip} \left( \frac{\pi_\theta(o_{i,t} | q, o_{i,<t})}{\pi_{\theta_{old}}(o_{i,t} | q, o_{i,<t})}, 1 - \epsilon, 1 + \epsilon \right)  \hat{A}_{i,t} \right] - \beta \mathbb{D}_{KL}\left[\pi_{\theta} || \pi_{ref}\right]\right\} ,
\end{split}
\label{eq:GRPO-obj}
\end{equation}
where $\epsilon$ and $\beta$ are hyper-parameters, and $\hat{A}_{i,t}$  is the advantage calculated based on relative rewards of the outputs inside each group only, which will be detailed in the following subsections. The group relative way that GRPO leverages to calculate the advantages, aligns well with the comparative nature of rewards models,  as reward models are typically trained on datasets of comparisons between outputs on the same question. Also note that, instead of adding KL penalty in the reward, GRPO regularizes by directly adding the KL divergence between the trained policy and the reference policy to the loss, avoiding complicating the calculation of  $\hat{A}_{i,t}$. And different from the KL penalty term used in (\ref{eq:PPO-reward}), we estimate the KL divergence with the following unbiased estimator \citep{kl_approx}: 
\begin{equation}
\small
    \mathbb{D}_{KL}\left[\pi_{\theta} || \pi_{ref}\right] = \frac{\pi_{ref}(o_{i,t}|q,o_{i,<t})}{\pi_{\theta}(o_{i,t}|q,o_{i,<t})}- \log\frac{\pi_{ref}(o_{i,t}|q,o_{i,<t})}{\pi_{\theta}(o_{i,t}|q,o_{i,<t})} - 1,
\end{equation}
which is guaranteed to be positive. 

% This section details our approach, which includes both outcome- and process-supervised GRPO.
\begin{algorithm}[t]
  \small
  \caption{Iterative Group Relative Policy Optimization}
  \textbf{Input} initial policy model $\pi_{\theta_{\text{init}}}$; reward models $r_\phi$; task prompts $\mathcal{D}$; 
  hyperparameters $\epsilon$, $\beta$, $\mu$
  \begin{algorithmic}[1]
    \State policy model $\pi_\theta \leftarrow \pi_{\theta_{\text{init}}}$
    \For{iteration = 1, \dots, I}
       \State reference model $\pi_{ref} \leftarrow \pi_{\theta}$
      \For{step = 1, \dots, M}
      \State Sample a batch $\mathcal{D}_b$ from $\mathcal{D}$
      \State Update the old policy model $\pi_{\theta_{old}} \leftarrow \pi_{\theta}$ 
      \State Sample $G$ outputs $\{o_i\}_{i=1}^G \sim \pi_{\theta_{old}} (\cdot \mid q) $ for each question $q \in \mathcal{D}_b$
      \State Compute rewards $\{r_i\}_{i=1}^{G}$ for each sampled output $o_i$ by running $r_{\phi}$ 
      \State Compute $\hat{A}_{i,t}$ for the $t$-th token of $o_i$ through group relative advantage estimation.
      \For{GRPO iteration = 1, \dots, $\mu$}
        \State Update the policy model $\pi_{\theta}$ by maximizing the GRPO objective (Equation \ref{eq:GC-GRPO})
      \EndFor
    \EndFor 
    \State Update $r_\phi$ through continuous training using a replay mechanism. 
    \EndFor 
  \end{algorithmic}
  \textbf{Output} $\pi_\theta$
  \label{alg:iter-grpo}
\end{algorithm}
\subsubsection{Outcome Supervision RL with GRPO} 
Formally, for each question $q$,  a group of outputs $\{o_1, o_2, \cdots, o_G\}$  are sampled from the old policy model $\pi_{\theta_{old}}$. A reward model is then used to score the outputs, yielding $G$ rewards  $\mathbf{r}=\{r_1, r_2, \cdots, r_G\}$ correspondingly. Subsequently, these rewards are normalized by subtracting the group average and dividing by the group standard deviation. Outcome supervision provides the normalized reward at the end of each output $o_i$  and sets the advantages  $\hat{A}_{i, t}$  of all tokens in the output as the normalized reward, i.e., $\hat{A}_{i, t} = \widetilde{r}_i = \frac{r_i- {\rm mean}(\mathbf{r})}{{\rm std}(\mathbf{r})}$, and then optimizes the policy by maximizing the objective defined in equation (\ref{eq:GRPO-obj}).

\subsubsection{Process Supervision RL with GRPO} 
Outcome supervision only provides a reward at the end of each output, which may not be sufficient and efficient to supervise the policy in complex mathematical tasks. Following \cite{wang2023math}, we also explore process supervision, which provides a reward at the end of each reasoning step. Formally, given the question $q$ and $G$  sampled outputs $\{o_1, o_2, \cdots, o_G\}$, a process reward model is used to score each step of the outputs, yielding corresponding rewards: $\mathbf{R} = \{ \{r_1^{index(1)},\cdots,r_1^{index(K_1)}\}, \cdots,  \{r_G^{index(1)},\cdots,r_G^{index(K_G)}\} \}$, where $index(j)$ is the end token index of the $j$-th step, and $K_i$ is the total number of steps in the $i$-th output. We also normalize these rewards with the average and the standard deviation, i.e., $\widetilde{r}_i^{index(j)} = \frac{r_i^{index(j)} - {\rm mean(\mathbf{R})}}{{\rm std(\mathbf{R})}}$.
Subsequently, the process supervision calculates the advantage of each token as the sum of the normalized rewards from the following steps, i.e., $\hat{A}_{i, t} = \sum_{index(j) \ge t} \widetilde{r}_i^{index(j)}$,
and then optimizes the policy by maximizing the objective defined in equation (\ref{eq:GRPO-obj}).

\subsubsection{Iterative RL with GRPO}
As the reinforcement learning training process progresses, the old reward model may not be sufficient to supervise the current policy model.
Therefore, we also explore the iterative RL with GRPO.
As shown in Algorithm \ref{alg:iter-grpo}, in iterative GRPO, we generate new training sets for the reward model based on the sampling results from the policy model and continually train the old reward model using a replay mechanism that incorporates 10\% of historical data.
Then, we set the reference model as the policy model, and continually train the policy model with the new reward model.

\subsection{Training and Evaluating \spmath-RL}

We conduct RL based on \spmath-Instruct 7B.
The training data of RL are chain-of-thought-format questions related to GSM8K and MATH from the SFT data, which consists of around 144K questions. 
We exclude other SFT questions to investigate the impact of RL on benchmarks that lack data throughout the RL phase.
We construct the training set of reward models following \citep{wang2023math}.
We train our initial reward model based on the \spmath-Base 7B with a learning rate of 2e-5.
For GRPO, we set the learning rate of the policy model as 1e-6. The KL coefficient is 0.04. For each question, we sample $64$ outputs.  The max length is set to 1024, and the training batch size is 1024.
The policy model only has a single update following each
exploration stage.
We evaluate \spmath-RL 7B on benchmarks following \spmath-Instruct 7B.
For \spmath-RL 7B, GSM8K and MATH with chain-of-thought reasoning can be regarded as in-domain tasks and all the other benchmarks can be regarded as out-of-domain tasks.

Table \ref{tab:sft_rl_math} demonstrates the performance of open- and closed-source models with both chain-of-thought and tool-integrated reasoning on English and Chinese benchmarks. We find that:
1) \spmath-RL 7B attains accuracies of 88.2\% and 51.7\% on GSM8K and MATH, respectively, utilizing chain-of-thought reasoning. This performance surpasses that of all open-source models in the 7B to 70B range, as well as the majority of closed-source models. 
2) Crucially, \spmath-RL 7B is only trained on chain-of-thought-format instruction tuning data of GSM8K and MATH, starting from \spmath-Instruct 7B. Despite the constrained scope of its training data, it outperforms \spmath-Instruct 7B across all evaluation metrics, showcasing the effectiveness of reinforcement learning.

\section{Discussion}
In this section, we will share our findings in pre-training and RL experiments. 

\subsection{Lessons Learnt in Pre-Training}

We first share our experience in pre-training. Unless otherwise specified, we will adhere to the training settings outlined in Section \ref{sec:quality-policy}. It is worth noting that, when referring to the DeepSeekMath Corpus in this section, we use an 89B-token dataset from the second iteration of the data collection process.

\subsubsection{Code Training Benefits Mathematical Reasoning}

A popular yet unverified hypothesis suggests that code training improves reasoning.
We attempt to offer a partial response to this, particularly within the mathematical domain:
code training improves models' ability to do mathematical reasoning both with and without tool use.

To study how code training affects mathematical reasoning, we experimented with the following two-stage training and one-stage training settings:

\noindent
\textbf{Two-Stage Training} 
\begin{itemize}[topsep=0pt]
    \item \textbf{Code Training for 400B Tokens $\rightarrow$ Math Training for 150B Tokens}:
    We train DeepSeek-LLM 1.3B for 400B code tokens followed by 150B math tokens;
    \item \textbf{General Training for 400B Tokens $\rightarrow$ Math Training for 150B Tokens}:
    As a control experiment, we also experiment with general tokens (sampled from a large-scale general corpus created by DeepSeek-AI) instead of code tokens in the first stage of training, in an attempt to investigate the advantages of code tokens over general tokens in improving mathematical reasoning.
\end{itemize}

\noindent
\textbf{One-Stage Training}
\begin{itemize}[topsep=0pt]
    \item \textbf{Math Training for 150B Tokens}:
    We train DeepSeek-LLM 1.3B for 150B math tokens;
    \item \textbf{Training on a mixture of 400B Code Tokens and 150B Math Tokens}:
    Math training following code training degrades coding performance.
    We investigate whether code tokens, when mixed with math tokens for one-stage training, would still improve mathematical reasoning and also alleviate the problem of catastrophic forgetting.
\end{itemize}

\begingroup
\setlength{\tabcolsep}{3pt} % Default value: 6pt
\renewcommand{\arraystretch}{1} % Default value: 1
\begin{table*}[t!]
    \centering
    \adjustbox{max width=\textwidth}{
\begin{tabular}{llllccccc} 
\toprule
\multirow{2}{*}{Training Setting} & \multicolumn{3}{l}{Training Tokens} & \multicolumn{3}{c}{w/o Tool Use} & \multicolumn{2}{c}{w/ Tool Use}  \\ 
\cmidrule(lr){2-4}\cmidrule(lr){5-7}\cmidrule(lr){8-9}
                          & General & Code & Math               & GSM8K  & MATH   & CMATH          & GSM8K+Python & MATH+Python       \\ 
\midrule
No Continual Training     & --                              & --                           & --                           & 2.9\%  & 3.0\%  & 12.3\%         & 2.7\%        & 2.3\%             \\ 
\midrule
\midrule
\multicolumn{9}{c}{Two-Stage Training}                                                                                                                                                          \\ 
% \midrule
% Stage 1: General Training (w/o LR Decay) & 400B                            & --                           & --                           & 2.4\%  & 2.8\%  & 12.3\%         & 1.7\%        & 1.8\%             \\
% Stage 2: Math Training    & --                              & --                           & 150B                         &    19.6\%    &     14.0\%   &        37.4\%        &    13.6\%          &     7.2\%           \\ 
\midrule
Stage 1: General Training & 400B                            & --                           & --                           & 2.9\%  & 3.2\%  & 14.8\%         & 3.3\%        & 2.3\%             \\
Stage 2: Math Training    & --                              & --                           & 150B                         &    19.1\%    &     14.4\%   &        37.2\%        &    14.3\%          &     6.7\%           \\ 
% \midrule
% Stage 1: Code Training (w/o LR Decay)    & --                              & 400B                         & --                           & 3.1\%  & 3.6\%  & 14.6\%         & 6.5\%        & 6.3\%             \\
% Stage 2: Math Training    & --                              & --                           & 150B                         & \textbf{22.2\%}  &  \textbf{14.8\%}  & \textbf{40.5\%}         & 18.7\%       & 9.1\%             \\ 
\midrule
Stage 1: Code Training    & --                              & 400B                         & --                           & 5.9\%  & 3.6\%  & 19.9\%         & 12.4\%        & 10.0\%             \\
Stage 2: Math Training    & --                              & --                           & 150B                         & \textbf{21.9\%}  &  \textbf{15.3\%}  & \textbf{39.7\%}         & 17.4\%       & 9.4\%             \\ 
\midrule
\midrule
\multicolumn{9}{c}{One-Stage Training}                                                                                                                                                          \\ 
\midrule
Math Training             & --                              & --                           & 150B                         & 20.5\% & 13.1\% & 37.6\%         & 11.4\%       & 6.5\%             \\
\midrule
Code \& Math Mixed Training & --                              & 400B                         & 150B                         &     17.6\%   &   12.1\%     &    36.3\%            &      \textbf{19.7\%}      &    \textbf{13.5\%}            \\
\bottomrule
\end{tabular}
    }
    \caption{
    Investigation of how code affects mathematical reasoning under different training settings.
    We experiment with DeepSeek-LLM 1.3B, and evaluate its mathematical reasoning performance without and with tool use via few-shot chain-of-thought prompting and few-shot program-of-thought prompting, respectively.
    }
    \label{tab:code-math}
\end{table*}
\endgroup

\begingroup
\setlength{\tabcolsep}{3pt} % Default value: 6pt
\renewcommand{\arraystretch}{1} % Default value: 1
\begin{table*}[t!]
    \centering
    \adjustbox{max width=\textwidth}{
\begin{tabular}{llllcccc} 
\toprule
\multirow{2}{*}{Training Setting} & \multicolumn{3}{l}{Training Tokens} & \multirow{2}{*}{MMLU} & \multirow{2}{*}{BBH} & \multirow{2}{*}{HumanEval (Pass@1)} & \multirow{2}{*}{MBPP (Pass@1)} \\ \cmidrule(lr){2-4}
& General & Code & Math &   &    &  &   \\ 
\midrule
No Continual Training & -- & -- & --     & 24.5\% & 28.1\% & 12.2\%             & 13.0\%         \\ 
\midrule
\multicolumn{8}{c}{Two-Stage Training}                                            \\ 
% \midrule
% Stage 1: General Training (w/o LR Decay)    & 27.5\% & 27.3\% & 9.8\%              & 13.2\%         \\
% Stage 2: Math Training             & 32.8\% & 33.5\% & 8.5\%              & 13.4\%         \\
\midrule
Stage 1: General Training & 400B & -- & --     & 25.9\% & 27.7\% & 15.2\%              & 13.6\%         \\
Stage 2: Math Training & -- & -- & 150B             & 33.1\% & 32.7\% & 12.8\%              & 13.2\%         \\
% \midrule
% Stage 1: Code Training (w/o LR Decay)            & 23.6\% & 29.8\% & 21.3\%             & 35.6\%         \\
% Stage 2: Math Training             & 35.3\% & 35.9\% & 14.0\%             & 19.8\%         \\
\midrule
Stage 1: Code Training & -- & 400B & --          & 25.0\% & 31.5\% & 25.0\%             & \textbf{40.0\%}         \\
Stage 2: Math Training & -- & -- & 150B             & \textbf{36.2\%} & 35.3\% & 12.2\%             & 17.0\%         \\
\midrule
\multicolumn{8}{c}{One-Stage Training}                                            \\ 
\midrule
Math Training & -- & -- & 150B             & 32.3\% & 32.5\% & 11.6\%             & 13.2\%         \\
\midrule
Code \& Math Mixed Training & -- & 400B & 150B & 33.5\% & \textbf{35.6\%} & \textbf{29.3\%}             & 39.4\%         \\
\bottomrule
\end{tabular}
    }
    \caption{
    Investigation of how different settings of code and math training affect model performance of language understanding, reasoning, and coding.
    We experiment with DeepSeek-LLM 1.3B.
    We evaluate the models on MMLU and BBH using few-shot chain-of-thought prompting.
    On HumanEval and MBPP, we conduct zero-shot and few-shot evaluations, respectively.
    }
    \label{tab:code-math-general-eval}
    \vspace{-0.2in}
\end{table*}
\endgroup

\paragraph{Results}
Table \ref{tab:code-math} and Table \ref{tab:code-math-general-eval} demonstrate the downstream performance under different training settings.

Code training benefits program-aided mathematical reasoning, both under the two-stage training and one-stage training settings.
As shown in Table \ref{tab:code-math}, under the two-stage training setting, code training alone already significantly enhances the ability to solve GSM8K and MATH problems using Python.
Math training in the second stage yields further improvements.
Interestingly, under the one-stage training setting, mixing code tokens and math tokens effectively mitigates the issue of catastrophic forgetting that arises from two-stage training, and also synergizes coding (Table \ref{tab:code-math-general-eval}) and program-aided mathematical reasoning (Table \ref{tab:code-math}).

Code training also improves mathematical reasoning without tool use.
Under the two-stage training setting, the initial stage of code training already results in moderate enhancements.
It also boosts the efficiency of the subsequent math training, eventually leading to the best performance.
However, combining code tokens and math tokens for one-stage training compromises mathematical reasoning without tool use.
One conjecture is that DeepSeek-LLM 1.3B, due to its limited scale, lacks the capacity to fully assimilate both code and mathematical data simultaneously.

\begingroup
\setlength{\tabcolsep}{3pt} % Default value: 6pt
\renewcommand{\arraystretch}{1} % Default value: 1
\begin{table*}[t!]
    \centering
    \adjustbox{max width=\textwidth}{
\begin{tabular}{lllcccccccc} 
\toprule
\multicolumn{1}{l}{\multirow{3}{*}{Model}} & \multicolumn{1}{l}{\multirow{3}{*}{Size}}      & \multirow{3}{*}{ArXiv Corpus} & \multicolumn{5}{c}{English Benchmarks}             & \multicolumn{3}{c}{Chinese Benchmarks}     \\ 
\cmidrule(lr){4-8}\cmidrule(lr){9-11}
 & &                              & GSM8K  & MATH   & OCW & SAT    & \makecell{MMLU \\ STEM} & CMATH  & \makecell{Gaokao \\ MathCloze} & \makecell{Gaokao \\ MathQA}  \\ 
\midrule
\multirow{3}{*}{DeepSeek-LLM} & \multirow{3}{*}{1.3B}   & No Math Training                                       & 2.9\%  & 3.0\%  & 2.9\%       & 15.6\% & 19.5\%    & 12.3\% & 0.8\%            & 17.9\%         \\
\cmidrule(lr){3-11}
& & MathPile                                             & 2.7\%  & 3.3\%  & 2.2\%       & 12.5\% & 15.7\%    & 1.2\%  & 0.0\%            & 2.8\%          \\
& & ArXiv-RedPajama                                     &    3.3\%    &  3.4\%      &   4.0\%        &   9.4\%    &      9.0\%   &   7.4\%   &          0.8\%      &  2.3\%             \\ 
\midrule
\multirow{3}{*}{DeepSeek-Coder-Base-v1.5} & \multirow{3}{*}{7B} & No Math Training                                       & 29.0\% & 12.5\% & 6.6\%       & 40.6\% & 38.1\%    & 45.9\% & 5.9\%            & 21.1\%         \\
\cmidrule(lr){3-11}
& & MathPile                                             & 23.6\% & 11.5\% & 7.0\%       & 46.9\% & 35.8\%    & 37.9\% & 4.2\%            & 25.6\%         \\
& & ArXiv-RedPajama                                     & 28.1\% & 11.1\% & 7.7\%       & 50.0\% & 35.2\%    & 42.6\% & 7.6\%            & 24.8\%         \\
\bottomrule
\end{tabular}
    }
    \caption{
    Effect of math training on different arXiv datasets.
    Model performance is evaluated with few-shot chain-of-thought prompting.
    }
    \label{tab:arxiv-cot}
\end{table*}
\endgroup

\begingroup
\setlength{\tabcolsep}{3pt} % Default value: 6pt
\renewcommand{\arraystretch}{1} % Default value: 1

\begin{table*}[t!]
    \centering
\begin{small}
\begin{tabular}{lcc} 
\toprule
ArXiv Corpus      & miniF2F-valid & miniF2F-test  \\ 
\midrule
No Math Training &   20.1\%        & 21.7\%            \\ 
\midrule
MathPile         & 16.8\%        & 16.4\%        \\
ArXiv-RedPajama  & 14.8\%        & 11.9\%        \\
\bottomrule
\end{tabular}
\end{small}
    \caption{
    Effect of math training on different arXiv corpora, the base model being DeepSeek-Coder-Base-v1.5 7B.
    We evaluate informal-to-formal proving in Isabelle.
    }
    \label{tab:arxiv_atp}
    \vspace{-0.1in}
\end{table*}
\endgroup

\subsubsection{ArXiv Papers Seem Ineffective in Improving Mathematical Reasoning}

ArXiv papers are commonly included as a component of math pre-training data \citep{minerva,gpt-f,llemma,mathpile}.
However, detailed analysis regarding their impact on mathematical reasoning has not been extensively conducted.
Perhaps counter-intuitively, according to our experiments, arXiv papers seem ineffective in improving mathematical reasoning.
We experiment with models of different sizes, including DeepSeek-LLM 1.3B and DeepSeek-Coder-Base-v1.5 7B \citep{deepseek-coder}, using arXiv corpora that underwent varied processing pipelines:
\begin{itemize}[topsep=0pt]
    \item \textbf{MathPile} \citep{mathpile}:
    an 8.9B-token corpus developed with cleaning and filtering heuristic rules, over 85\% of which are scientific arXiv papers;
    \item \textbf{ArXiv-RedPajama} \citep{redpajama}:
    the entirety of arXiv LaTeX files with preambles, comments, macros, and bibliographies removed, totaling 28.0B tokens.
\end{itemize}
\noindent
In our experiments, we separately train DeepSeek-LLM 1.3B for 150B tokens and DeepSeek-Coder-Base-v1.5 7B for 40B tokens on each arXiv corpus. It seems that arXiv papers are ineffective in improving mathematical reasoning.
When trained on a arXiv-only corpus, both models display no notable improvements or even deterioration across various mathematical benchmarks of different complexities employed in this study.
These benchmarks include quantitative reasoning datasets like GSM8K and MATH (Table \ref{tab:arxiv-cot}), multiple-choice challenges like MMLU-STEM (Table \ref{tab:arxiv-cot}), and formal mathematics like miniF2F (Table \ref{tab:arxiv_atp}).

However, this conclusion has its limitations and should be taken with a grain of salt.
We have not yet studied:
\begin{itemize}[topsep=0pt]
    \item The impact of arXiv tokens on specific math-related tasks not included in this research, such as informalization of theorems which is to convert formal statements or proofs to their informal versions;
    \item The effect of arXiv tokens when combined with other types of data;
    \item Whether the benefits of arXiv papers would manifest themselves at a larger model scale.
\end{itemize}
Thus, further exploration is required, which we leave for future studies.

\subsection{Insights of Reinforcement Learning}
\subsubsection{Towards to a Unified Paradigm}
In this section, we provide a unified paradigm to analyze different training methods, such as SFT, RFT, DPO, PPO, GRPO, and further conduct experiments to explore the factors of the unified paradigm. 
Generally, the gradient with respect to the parameter $\theta$ of a training method can be written as:
\begin{equation}
    \nabla_{\theta}\mathcal{J}_{\textcolor{red}{\mathcal{A}}}(\theta) = \mathbb{E}[\underbrace{(q,o) \sim \textcolor{red}{\mathcal{D}}}_{Data \ Source}]\left( \frac{1}{|o|} \sum_{t=1}^{|o|}  \underbrace{GC_{{\mathcal{A}}}(q, o, t, \textcolor{red}{\pi_{{rf}}})}_{Gradient \ Coefficient}  \nabla_{\theta}\log \pi_{\theta}(o_t | q, o_{<t})\right).
\label{eq:objective}
\end{equation}
There exist three key components: 
1) \textit{Data Source $\mathcal{D}$}, which determines the training data;
2) \textit{Reward Function $\pi_{{rf}}$}, which is the source of the training reward signal;
3) \textit{Algorithm $\mathcal{A}$}: which processes the training data and the reward signal to the gradient coefficient $GC$ that determines the magnitude of the penalty or reinforcement for the data. We analyze several representative methods based on such a unified paradigm:

\begin{table*}[t]
    \centering
    \begin{small}
    \begin{tabular}{lccc}
    \toprule
     \textbf{Methods} & \textbf{Data Source}  & \textbf{Reward Function} & \textbf{Gradient Coefficient}  \\
    \midrule
    SFT & $q, o \sim P_{sft}(Q, O)$& - &  1 \\
    \midrule
    RFT  & $q \sim P_{sft}(Q)$, $o \sim \pi_{sft}(O|q)$ & Rule & Equation \ref{eq:GC-RFT} \\
     DPO & $q \sim P_{sft}(Q)$, $o^+, o^- \sim \pi_{sft}(O|q)$ & Rule & Equation \ref{eq:GC-DPO} \\
    \midrule
    Online RFT  & $q \sim P_{sft}(Q)$, $o \sim \pi_{\theta}(O|q)$ & Rule & Equation \ref{eq:GC-RFT} \\
    PPO & $q \sim P_{sft}(Q)$, $o \sim \pi_{\theta}(O|q)$ & Model & Equation \ref{eq:GC-PPO} \\
    GRPO & $q \sim P_{sft}(Q)$, $\{o_i\}_{i=1}^G \sim \pi_{\theta}(O|q)$ & Model & Equation \ref{eq:GC-GRPO} \\
    % PPOwoC+PS & $q \sim P_{sft}(Q)$, $o \sim \pi_{\theta}(O|q)$ & $A_{prm} + KL$ (Equation \ref{eq:GC-PPO}) \\
    \bottomrule
    \end{tabular}
    \end{small}
    \caption{The data source and gradient coefficient of different methods. $P_{sft}$ denotes the data distribution of supervised fine-tuning datasets. $\pi_{\theta_{sft}}$ and $\pi_{\theta}$ denote the supervised fine-tuned model and the real-time policy model during the online training process, respectively.}
    \label{tab:data_policy}
    % \vspace{-0.1in}
\end{table*}

\begin{itemize}
    \item  \textbf{Supervised Fine-tuning (SFT)}: SFT fine-tunes pretrained model on human selected SFT data.
    \item \textbf{Rejection Sampling Fine-tuning (RFT)}: RFT further fine-tunes the SFT model on the filtered outputs sampled from the SFT model based on SFT questions. RFT filters the outputs based on the correctness of their answers.
    \item \textbf{Direct Preference Optimization (DPO)}: DPO further refines the SFT model by fine-tuning it on augmented outputs sampled from the SFT model, using pair-wise DPO loss.
    \item \textbf{Online Rejection Sampling Fine-tuning (Online RFT)}: Different from RFT, Online RFT initiates the policy model using the SFT model and refines it by fine-tuning with the augmented outputs sampled from the real-time policy model.
    \item \textbf{PPO/GRPO}: PPO/GRPO initializes the policy model using the SFT model and reinforces it with the outputs sampled from the real-time policy model.
\end{itemize}
We summarize the components of these methods in Table \ref{tab:data_policy}.
Please refer to Appendix \ref{app:analysis-rl} for a more detailed derivation process.

\begin{figure*}[t]
\centering
\includegraphics[width=0.9\linewidth]{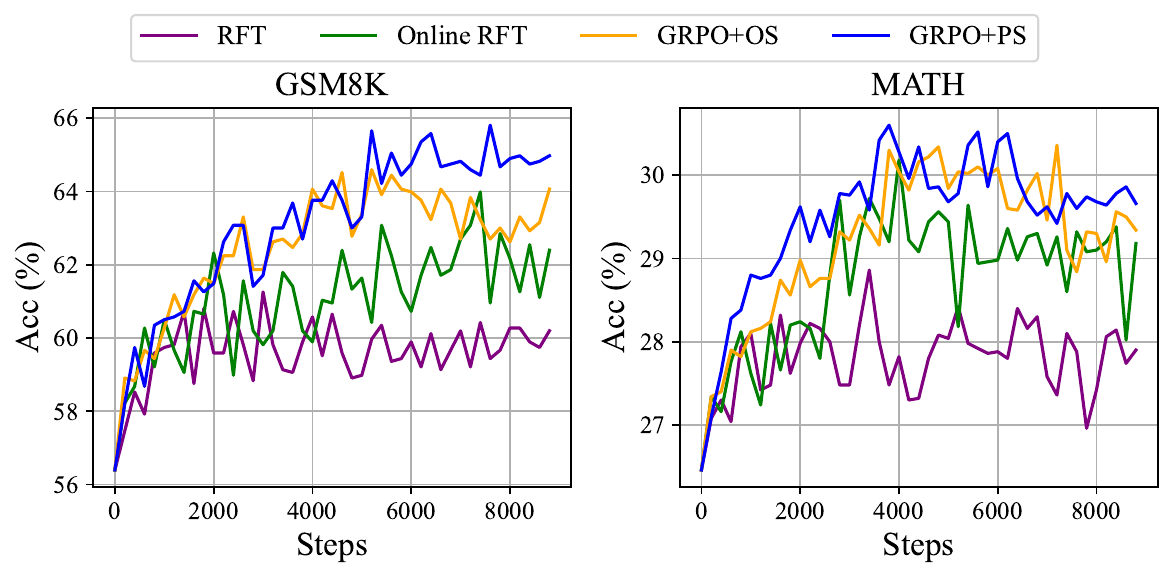}\vspace{-0.1in}
\caption{Performance of the \spmath-Instruct 1.3B model, which was further trained using various methods, on two benchmarks.}
\label{fig:rl-analysis}
\end{figure*}

\begin{figure*}[t]
\centering
\includegraphics[width=0.9\linewidth]{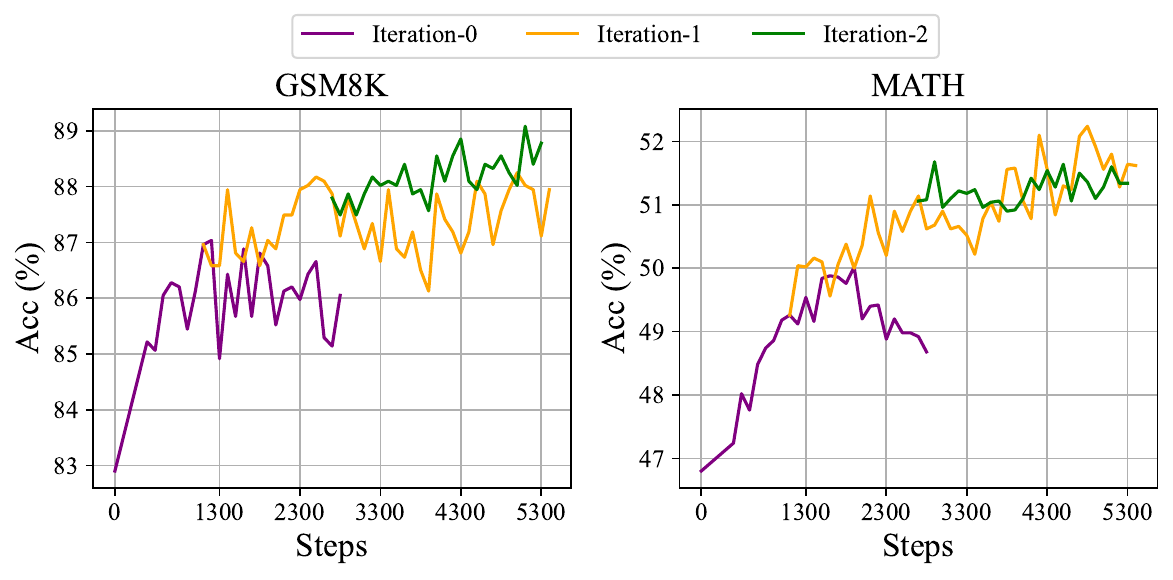}\vspace{-0.1in}
\caption{Performance of iterative reinforcement learning with \spmath-Instruct 7B on two benchmarks.}
\label{fig:iter-rl}
\end{figure*}

\paragraph{Observation about Data Source} 
We divide the data source into two categories, online sampling, and offline sampling.
Online sampling denotes that the training data is from the exploration results of the real-time training policy model, while offline sampling denotes that the training data is from the sampling results of the initial SFT model.
RFT and DPO follow the offline style, while Online RFT and GRPO follow the online style. 

As shown in Figure \ref{fig:rl-analysis},
we find that the Online RFT significantly outperforms RFT on two benchmarks.
Specifically, Online RFT is comparable to RFT in the early stage of training but gains an absolute advantage in the later stage, demonstrating the superiority of online training. 
This is intuitive, as in the initial stage, the actor and the SFT model exhibit close resemblance, with the sampled data revealing only minor differences. In the later stage, however, the data sampled from the actor will exhibit more significant differences, and real-time data sampling will offer greater advantages.

\paragraph{Observation about Gradient Coefficient}
The algorithm processes the reward signal to the gradient coefficient to update the model parameter.
We divide the reward function as `Rule' and `Model' in our experiments.
Rule refers to judging the quality of a response based on the correctness of the answer, and Model denotes that we train a reward model to score each response. The training data of the reward model is based on the rule judgment.
Equations \ref{eq:GC-RFT} and \ref{eq:GC-GRPO} highlight a key difference between GRPO and Online RFT: GRPO uniquely adjusts its gradient coefficient based on the reward value provided by the reward model. This allows for differential reinforcement and penalization of responses according to their varying magnitudes. In contrast, Online RFT lacks this feature; it does not penalize incorrect responses and uniformly reinforces all responses with correct answers at the same level of intensity. 

As demonstrated in Figure \ref{fig:rl-analysis}, GRPO surpasses online RFT, thereby highlighting the efficiency of altering positive and negative gradient coefficients. In addition, GRPO+PS shows superior performance compared to GRPO+OS, indicating the benefits of using fine-grained, step-aware gradient coefficients.
Furthermore, we explore the iterative RL, in our experiments, we conduct two rounds of iteration. As shown in Figure \ref{fig:iter-rl}, we notice that the iterative RL significantly improves the performance, especially at the first iteration.

\begin{figure}[t]
\centering
\includegraphics[width=0.95\linewidth]{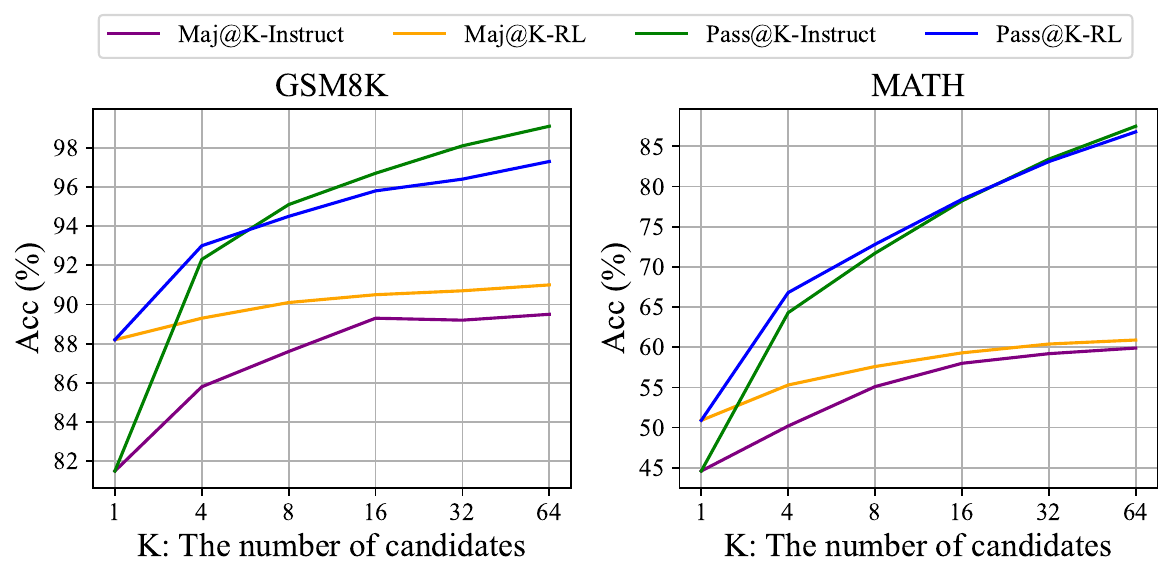}\vspace{-0.1in}
\caption{The Maj@K and Pass@K of SFT and RL \spmath~7B on GSM8K and MATH (temperature $0.7$).  It was noted that RL enhances Maj@K but not Pass@K.}
\label{fig:maj-pass}
\end{figure}

\subsubsection{Why RL Works?}
In this paper, we conduct reinforcement learning based on a subset of instruction tuning data, and it achieves significant performance enhancement upon the instruction tuning model. To further explain why reinforcement learning works.
We evaluate the Pass@K and Maj@K accuracy of the Instruct and RL models on two benchmarks.
As shown in Figure \ref{fig:maj-pass}, RL enhances Maj@K’s performance but not Pass@K. These findings indicate that RL enhances the model's overall performance by rendering the output distribution more robust, in other words, \textbf{it seems that the improvement is attributed to boosting the correct response from TopK rather than the enhancement of fundamental capabilities.} Similarly, \citep{wang2023making} identified a \textbf{misalignment problem} in reasoning tasks within the SFT model, showing that the reasoning performance of SFT models can be improved through a series of preference alignment strategies \citep{yuan2023rrhf,song2023preference,wang2023making}.

\subsubsection{How to Achieve More Effective RL?}
\label{sec:effec_rl}
We demonstrate RL works pretty well in mathematical reasoning tasks. We also provide a unified paradigm to understand different representative training methods.
Within this paradigm, all methods are conceptualized as either direct or simplified  RL techniques. 
As summarized in Equation \ref{eq:objective}, there exist three key components: Data Source, Algorithm, and Reward Function. We provide some potential future directions about the three components.

\paragraph{Data Source} Data source is the raw material of all training methods.
In the context of RL, we specifically refer to the data source as the unlabeled questions with the outputs sampled from the policy model. In this paper, we only use the questions from the instruction tuning stage and a naive nucleus sampling to sample outputs. We think this is a potential reason that 
our RL pipeline only improves the Maj@K performance. In the future, we will explore our RL pipeline on out-of-distribution question prompts, in conjunction with \textbf{advanced sampling (decoding) strategies}, like those based on tree-search methods \citep{yao2023tree}.
Also, the  \textbf{efficient inference techniques} \citep{xia-etal-2023-speculative,leviathan2023fast,kwon2023efficient,xia2024unlocking}, which determines the exploration efficiency of policy models, also play an exceedingly important role.

\paragraph{Algorithms} Algorithms process the data and reward signal to the gradient coefficient to update the model parameter. 
Based on Equation \ref{eq:objective}, to some extent, all methods now fully \textbf{TRUST} the signal of the reward function to increase or decrease the conditional probability of a certain token.
However, it is impossible to ensure the reward signal is always reliable, especially in extremely complex tasks. For example, even the PRM800K datasets \citep{lightman2023let}, which have been carefully annotated by well-trained annotators, still contain approximately 20\% of incorrectly annotations\footnote{\url{https://github.com/openai/prm800k/issues/12\#issuecomment-1728491852}}. To this end, we will explore the reinforcement learning algorithm that is robust against noisy reward signals.  We believe such \textbf{WEAK-TO-STRONG} \citep{burns2023weak} alignment methods will bring a fundamental change to the learning algorithms.

\paragraph{Reward Function} Reward function is the source of the training signal.
In RL, the reward function is usually the neural reward model.
We think there exist three important directions for reward models:
1) \textbf{How to enhance the generalization ability of the reward model.} The reward model must be effectively generalized to handle out-of-distribution questions and advanced decoding outputs; otherwise, reinforcement learning may merely stabilize the distribution of LLMs rather than improve their fundamental capabilities;
2) \textbf{How to reflect the uncertainty of reward model.} The uncertainty could potentially act as a linking bridge between the weak reward model and the weak-to-strong learning algorithms;
3) \textbf{How to efficiently build high-quality process reward models} that can provide fine-grained training signals for the reasoning process \citep{lightman2023let,wang2023math}.

\section{Conclusion, Limitation, and Future Work}
We present \spmath, which outperforms all open-source models on the competition-level MATH benchmark and approaches the performance of closed models.
\spmath~is initialized with DeepSeek-Coder-v1.5 7B and undergoes continual training for 500B tokens, with a significant component of the training data being 120B math tokens sourced from Common Crawl.
Our extensive ablation study shows web pages offer significant potential for high-quality mathematical data, while arXiv may not as beneficial as we expected.
We introduce Group Relative Policy Optimization (GRPO), a variant of Proximal Policy Optimization (PPO), which can notably improve mathematical reasoning capabilities with less memory consumption.
The experiment results show that GRPO is effective even if \spmath-Instruct 7B has reached a high score on benchmarks. 
We also provide a unified paradigm to understand a series of methods and summarize several potential directions for more effective reinforcement learning.

Although \spmath~achieves impressive scores on quantitative reasoning benchmarks, its capability on geometry and theorem-proof are relatively weaker than closed models.
For instance, in our dry run, the model cannot handle problems related to triangles and ellipses, which may indicate data selection bias in pre-training and fine-tuning. In addition, restricted by the model scale, \spmath~is worse than GPT-4 on few-shot capability.
GPT-4 could improve its performance with few-shot inputs, while \spmath~shows similar performance in zero-shot and few-shot evaluation.
In the future, we will further improve our engineered data selection pipeline to construct more high-quality pre-trained corpus.
In addition, we will explore the potential directions (Section \ref{sec:effec_rl}) for more effective reinforcement learning of LLMs.

\newpage

\bibliography{main}

\newpage
\appendix

\section{Appendix}
\subsection{Analysis of Reinforcement Learning}
\label{app:analysis-rl}
We provide the detailed derivation of the data source and gradient coefficient (algorithm and reward function) across various methods, including SFT, RFT, Online RFT, DPO, PPO, and GRPO.

\subsubsection{Supervised Fine-tuning}
The objective of Supervised Fine-tuning is maximizing the following objective:
\begin{equation}
\mathcal{J}_{SFT}(\theta)=\mathbb{E}[q, o \sim P_{sft}(Q, O)]\left(\frac{1}{|o|}\sum_{t=1}^{|o|} \log \pi_\theta(o_t | q, o_{<t})\right).
\label{eq:mle}
\end{equation}
The gradient of $\mathcal{J}_{SFT}(\theta)$ is:
\begin{equation}
\nabla_{\theta}\mathcal{J}_{SFT} = \mathbb{E}[q, o \sim P_{sft}(Q, O)]\left(\frac{1}{|o|}\sum_{t=1}^{|o|} \nabla_{\theta} \log \pi_\theta(o_{t} | q, o_{<t})\right).
\end{equation}
Data Source: The dataset employed for SFT. Reward Function: This can be regarded as human selection. Gradient Coefficient: always set to 1.

\subsubsection{Rejection Sampling Fine-tuning} 
Rejection Sampling Fine-tuning first samples multiple outputs from the supervised fine-tuned LLMs for each question, and then trains LLMs on the sampled outputs with the correct answer.
Formally, the objective of RFT is to maximize the following objectives:
\begin{equation}
\mathcal{J}_{RFT}(\theta)= \mathbb{E}[q \sim P_{sft}(Q), o \sim \pi_{sft}(O|q)]\left( \frac{1}{|o|}\sum_{t=1}^{|o|} \mathbb{I}(o) \log \pi_\theta(o_{t} | q, o_{<t})\right).
\end{equation}
The gradient of $\mathcal{J}_{RFT}(\theta)$ is:
\begin{equation}
\nabla_{\theta}\mathcal{J}_{RFT}(\theta)= \mathbb{E}[{q \sim P_{sft}(Q), o \sim \pi_{sft}(O|q)}]\left( \frac{1}{|o|}\sum_{t=1}^{|o|} {\mathbb{I}(o)} \nabla_{\theta}\log \pi_\theta(o_{t} | q, o_{<t})\right).
\end{equation}
Data Source: question in SFT dataset with outputs sampled from SFT model. Reward Function: Rule (whether the answer is correct or not). Gradient Coefficient:
\begin{equation}
GC_{RFT}(q, o, t) = \mathbb{I}(o)=\left\{
\begin{aligned}
1  & & {\rm the \ answer \ of \ o \ is \ correct} \\
0  & & {\rm the \ answer \ of \ o \ is \ incorrect} \\
\end{aligned}
\right.
\label{eq:GC-RFT}
\end{equation}
\subsubsection{Online Rejection Sampling Fine-tuning}
The only difference between RFT and Online RFT is that the outputs of Online RFT are sampled from the real-time policy model $\pi_{\theta}$, rather than from the SFT model $\pi_{\theta_{sft}}$. Therefore, the gradient of online RFT is:
\begin{equation}
\nabla_{\theta}\mathcal{J}_{OnRFT}(\theta)= \mathbb{E}[{q \sim P_{sft}(Q), o \sim \pi_{\theta}(O|q)}]\left( \frac{1}{|o|}\sum_{t=1}^{|o|} {\mathbb{I}(o)} \nabla_{\theta}\log \pi_\theta(o_{t} | q, o_{<t})\right).
\end{equation}

\subsubsection{Direct Preference Optimization (DPO)}
The objective of DPO is:
\begin{equation}
\footnotesize
\begin{split}
    \mathcal{J}_{DPO}(\theta) = \mathbb{E}{[q \sim P_{sft}(Q), o^+, o^- \sim \pi_{sft}(O|q)]} \log \sigma \left(  \beta \frac{1}{|o^+|}\sum_{t=1}^{|o^+|} \log \frac{\pi_{\theta}(o^+_t | q, o^+_{<t})}{\pi_{\text{ref}}(o^+_t | q, o^+_{<t})} - \beta \frac{1}{|o^-|}\sum_{t=1}^{|o^-|} \log \frac{\pi_{\theta}(o^-_{<t} | q, o^-_{<t})}{\pi_{\text{ref}}(o^-_{<t} | q,o^-_{<t})} \right) 
\end{split}
\end{equation}
The gradient of $\mathcal{J}_{DPO}(\theta)$ is:
\begin{equation}
\footnotesize
\begin{split}
    \nabla_{\theta}\mathcal{J}_{DPO}(\theta)  = \mathbb{E}{[q \sim P_{sft}(Q), o^+, o^- \sim \pi_{sft}(O|q)]}
     & \left( \frac{1}{|o^+|}\sum_{t=1}^{|o^+|} GC_{DPO}  (q,o,t) \nabla_{\theta}\log\pi_{\theta}(o^+_t | q, o^+_{<t}) \right. \\
    - & \left. \frac{1}{|o^-|}\sum_{t=1}^{|o^-|}  GC_{DPO}  (q,o,t) \nabla_{\theta}\log\pi_{\theta}(o^-_t | q, o^-_{<t}) \right)
\end{split}
\end{equation}
Data Source:  question in SFT dataset with outputs sampled from SFT model.
Reward Function: human preference in the general domain (can be `Rule' in mathematical tasks).
Gradient Coefficient: 
\begin{equation}
\footnotesize
GC_{DPO}(q,o,t) = \sigma\left(\beta\log \frac{\pi_{\theta}(o^-_t | q, o^-_{<t})}{\pi_{\text{ref}}(o^-_t | q, o^-_{<t})} - \beta\log \frac{\pi_{\theta}(o^+_t | q, o^+_{<t})}{\pi_{\text{ref}}(o^+_t | q, o^+_{<t})}\right) 
\label{eq:GC-DPO}
\end{equation}

\subsubsection{Proximal Policy Optimization (PPO)}
The objective of PPO is:

\begin{equation}
\footnotesize
    \mathcal{J}_{PPO}(\theta) = \mathbb{E}{[q \sim P_{sft}(Q), o \sim \pi_{\theta_{old}}(O|q)]} \frac{1}{|o|} \sum_{t=1}^{|o|} \min \left[ \frac{\pi_\theta(o_{t} | q, o_{<t})}{\pi_{\theta_{old}}(o_{t} | q, o_{<t})} A_{t}, \text{clip} \left( \frac{\pi_\theta(o_{t} | q, o_{<t})}{\pi_{\theta_{old}}(o_{t} | q, o_{<t})}, 1 - \epsilon, 1 + \epsilon \right)  A_{t} \right].
\end{equation}
To simplify the analysis,  it is assumed that the model only has a single update following each exploration stage, thereby ensuring that $\pi_{\theta_{old}} = \pi_{\theta}$.
In this case, we can remove the $\min$ and ${\rm clip}$ operation:
\begin{equation}
\footnotesize
    \mathcal{J}_{PPO}(\theta) = \mathbb{E}{[q \sim P_{sft}(Q), o \sim \pi_{\theta_{old}}(O|q)]} \frac{1}{|o|} \sum_{t=1}^{|o|} \frac{\pi_\theta(o_{t} | q, o_{<t})}{\pi_{\theta_{old}}(o_{t} | q, o_{<t})} A_{t}.
\end{equation}
The gradient of $\mathcal{J}_{PPO}(\theta)$ is:
\begin{equation}
\footnotesize
\begin{split}
    \nabla_{\theta}\mathcal{J}_{PPO}(\theta) = \mathbb{E}{[q \sim P_{sft}(Q), o \sim \pi_{\theta_{old}}(O|q)]} \frac{1}{|o|} \sum_{t=1}^{|o|} A_t \nabla_{\theta}\log \pi_\theta(o_{t} | q, o_{<t})
\end{split}
\end{equation}
Data Source:  question in SFT dataset with outputs sampled from policy model.
Reward Function: reward model.
Gradient Coefficient:
\begin{equation}
    GC_{PPO}(q, o, t, \pi_{\theta_{rm}}) = A_t,
\label{eq:GC-PPO}
\end{equation}
where $A_t$ is the advantage, which is computed by applying Generalized Advantage Estimation (GAE) \citep{gae}, based on the rewards  $\{r_{\ge t}\}$  and a learned value function  $V_{\psi}$.

% \begin{equation}
%     GC_{PPO}(q, o, t, \pi_{\theta_{rm}}) = A_t(\pi_{\theta_{rm}}) + \beta \left( \frac{\pi_{ref}(o_{t}|o_{<t})}{\pi_{\theta}(o_{t}|o_{<t})}- 1 \right)
% \label{eq:GC-PPO}
% \end{equation}
\subsubsection{Group Relative Policy Optimization (GRPO)}
The objective of GRPO is (assume $\pi_{\theta_{old}} = \pi_{\theta}$ for simplified analysis):
\begin{equation}
\footnotesize
\begin{split}
    \mathcal{J}_{GRPO}(\theta) &= \mathbb{E}{[q \sim P_{sft}(Q), \{o_i\}_{i=1}^G \sim \pi_{\theta_{old}}(O|q)]}  \\
    & \frac{1}{G}\sum_{i=1}^G\frac{1}{|o_i|} \sum_{t=1}^{|o_i|} \left[\frac{\pi_\theta(o_{i,t} | q, o_{i,<t})}{\pi_{\theta_{old}}(o_{i,t} | q, o_{i,<t})} \hat{A}_{i,t} - \beta (\frac{\pi_{ref}(o_{i,t}|q,o_{i,<t})}{\pi_{\theta}(o_{i,t}|q,o_{i,<t})}- \log\frac{\pi_{ref}(o_{i,t}|q,o_{i,<t})}{\pi_{\theta}(o_{i,t}|q,o_{i,<t})} - 1)\right].
\end{split}
\end{equation}
The gradient of $\mathcal{J}_{GRPO}(\theta)$ is:
\begin{equation}
\footnotesize
\begin{split}
    \nabla_{\theta}\mathcal{J}_{GRPO}(\theta)  & = \mathbb{E}{[q \sim P_{sft}(Q), \{o_i\}_{i=1}^G \sim \pi_{\theta_{old}}(O|q)]} \\
    & \frac{1}{G}\sum_{i=1}^G\frac{1}{|o_i|} \sum_{t=1}^{|o_i|}  
    \left[\hat{A}_{i,t} + \beta \left(\frac{\pi_{ref}(o_{i,t}|o_{i,<t})}{\pi_{\theta}(o_{i,t}|o_{i,<t})} - 1\right)\right]  \nabla_{\theta}\log \pi_\theta(o_{i,t} | q, o_{i,<t}). 
\end{split}
\end{equation}
Data Source:  question in SFT dataset with outputs sampled from policy model.
Reward Function: reward model.
Gradient Coefficient:
\begin{equation}
\footnotesize
    GC_{GRPO}(q, o, t, \pi_{\theta_{rm}}) = \hat{A}_{i,t} + \beta \left(\frac{\pi_{ref}(o_{i,t}|o_{i,<t})}{\pi_{\theta}(o_{i,t}|o_{i,<t})} - 1\right),
\label{eq:GC-GRPO}
\end{equation}
where $\hat{A}_{i,t}$ is computed based on the group reward scores.

\end{CJK*}
\end{document}

%% file: commands.tex
\renewcommand{\phi}{\varphi}

\renewcommand{\epsilon}{\varepsilon}
\renewcommand{\imath}{\mathrm{i}}

\newlength{\restsubwidth}
\newlength{\restsubheight}
\newlength{\restsubmoreheight}
\newcommand{\rest}[2]{%
        \settowidth{\restsubwidth}{\ensuremath{#2}}
        \settoheight{\restsubheight}{\ensuremath{{}_{#2}}}
        \ensuremath{{#1\hskip 0.5pt}_{\vrule\kern2pt\parbox[b][%
        4pt][b]{\the\restsubwidth}{%
                        \ensuremath{{}_{#2}}}}}
        }